\documentclass{article}

\usepackage{PRIMEarxiv}

\usepackage[english]{babel}
\usepackage{microtype}
\usepackage{booktabs}
\usepackage{etoolbox}

\usepackage[T1]{fontenc}

\usepackage{graphicx}
\usepackage{diagbox}
\usepackage{amsmath}
\usepackage{amssymb}
\usepackage{amsthm}
\usepackage{subcaption}
\usepackage{hyperref}
\usepackage{siunitx}

\newcommand{\R}{\mathbb{R}}
\newcommand{\N}{\mathbb{N}}
\newcommand{\K}{\mathcal{K}}
\newcommand{\C}{\mathcal{C}}

\DeclareMathOperator{\PE}{PE}
\DeclareMathOperator{\LWPE}{LWPE}
\DeclareMathOperator{\MSE}{MSE}
\DeclareMathOperator{\RMSE}{RMSE}
\DeclareMathOperator{\MAE}{MAE}
\DeclareMathOperator{\ReLU}{ReLU}
\DeclareMathOperator{\BNN}{BNN}
\DeclareMathOperator{\CPLF}{CPLF}
\DeclareMathOperator{\step}{step}
\DeclareMathOperator{\reff}{ref}
\DeclareMathOperator{\pred}{pred}

\newtheorem{example}{Example}
 \newtheorem{thm}{Theorem}
 
 \newtheorem{lem}{Lemma}
 
 \newtheorem{defn}{Definition}
 \newtheorem{rem}{Remark}
 
\fancypagestyle{firstpage}{
  \fancyhf{} % limpia todo
}

\title{Barycentric Neural Networks and Length-Weighted Persistent Entropy Loss: A Green Geometric and Topological Framework for Function Approximation
}

\author{ \href{https://orcid.org/0009-0006-1316-9026}{\includegraphics[scale=0.06]{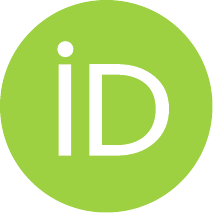}}Victor Toscano-Duran\thanks{Corresponding author.}, \href{https://orcid.org/0000-0001-9937-0033}{\includegraphics[scale=0.06]{orcid.pdf}}Rocio Gonzalez-Diaz \\
  Department of Applied Mathematics I, University of Seville \\
  Seville, Spain \\
  \texttt{\{vtoscano, rogodi\}us.es} \\
   \And
  \href{https://orcid.org/0000-0002-3624-6139}{\includegraphics[scale=0.06]{orcid.pdf}}Miguel A. Gutiérrez-Naranjo \\
  Department of Computer Science and Artificial Intelligence, University of Seville \\ Seville, Spain \\
  \texttt{magutier@us.es} \\
}

\begin{document}
\maketitle
\thispagestyle{firstpage}

\begin{abstract} 
While artificial neural networks are known as \emph{universal approximators} for continuous functions, many modern approaches rely on overparameterized architectures with high computational cost. In this work, we introduce the \emph{Barycentric Neural Network} ($\BNN$): a \emph{compact shallow architecture} that encodes both structure and parameters through a fixed set of \emph{base points} and their associated \emph{barycentric coordinates}. We show that the $\BNN$ enables the exact representation of \emph{continuous piecewise linear functions} ($\CPLF$s), ensuring strict continuity across segments. Given that any continuous function on a compact domain can be uniformly approximated by $\CPLF$s, the $\BNN$ emerges as a flexible and interpretable tool for \emph{function approximation}. 

To enhance geometric fidelity in low-resource scenarios, such as those with few base points to create $\BNN$s or limited training epochs, we propose \emph{length-weighted persistent entropy} ($\LWPE$): a stable variant of persistent entropy. Our approach integrates the $\BNN$ with a loss function based on $\LWPE$ to \emph{optimize the base points that define the $\BNN$}, rather than its internal parameters. Experimental results show that our approach achieves \emph{superior and faster approximation performance} compared to standard losses (MSE, RMSE, MAE and LogCosh), offering a computationally \emph{sustainable alternative for function approximation}.

\keywords{Barycentric Coordinates \and Neural Networks \and  Persistent Entropy \and  Function Approximation \and Green AI}

\end{abstract}

\section{Introduction}\label{sec:intro}

\emph{Function representation and approximation} are fundamental problems in machine learning and scientific computing, underpinning a wide range of applications such as physical simulation, data modeling, and predictive analysis. A cornerstone of this field is the \emph{Universal Approximation Theorem}, which guarantees that \emph{artificial neural networks} (ANNs) can approximate any continuous function on a compact domain \cite{HORNIK1989Multilayerfeedforwardnetworksareuniversalapproximators}. Since then, theoretical advances have deepened our understanding of ANN's approximation capabilities, particularly for smooth functions \cite{ferrari2005smooth,zainuddin2008functionapproximationann,liang2017deepneuralnetworksfunction}, and have highlighted the expressive power of ReLU-based architectures for representing piecewise linear structures \cite{LIU2021OptimalFunctionApproximationwithReluNN,PETERSEN2018OptimalApproximationofpiecewisesmoothfunctionsusingdeepreluneuralnetowkrs}. More recent work has also explored the approximation efficiency \cite{bolcskey2019OptimalApproximationwithSparseleyConnectedNeuralNetwork} and the role of depth versus width in function approximation \cite{Safran2017DepthWidthtradeoffsinapproximationfunctionwithneuralnetworks}. However, modern deep neural networks (DNNs) typically require highly overparameterized architectures, large datasets, and extensive computational resources. This leads to architectures (models) that are not only energy-intensive and opaque, but also poorly suited for resource-constrained environments, such as embedded systems or edge devices, where lightweight and efficient solutions are essential.

This tension between performance and sustainability has spurred interest in methods that can achieve high performance with reduced computational overhead. In particular, the emerging paradigm of Green AI \cite{bolon2024reviewgreenai} calls for machine learning methods that balance performance with computational and environmental cost. Within this context, \emph{geometric} tools like barycentric coordinates offer a promising yet underexplored avenue. These coordinates allow points to be expressed as convex combinations of the vertices of a simplex, enabling a natural mechanism for piecewise linear interpolation. Historically applied in computer graphics and geometric design \cite{warren2007barycentric}, barycentric interpolation has recently been extended to function approximation, providing interpretability and expressivity \cite{PALUZOHIDALGO2020TwohiddenlayerFeedforwardNetworkareUniversalApproximators,TrainableandExplainablePALUZOHIDALGO2024120474}. However, these methods still rely on barycentric subdivisions, which could limit flexibility. In contrast, our approach does not rely on subdivisions. In addition, nested barycentric coordinate systems (NBCS) have been proposed to represent complex function spaces in an interpretable manner \cite{Gottlieb2024NestedBarycentricCoordinateSystem}. Similarly, variational barycentric interpolation has been explored for mesh-based learning and shape optimization \cite{Dodik2023VariationalBarycentricCoordinates}.

Besides, \emph{topological} methods have also been integrated into data analysis and machine learning. \emph{Persistent homology}, a core method in \emph{topological data analysis} (TDA), and its associated summaries, like \emph{persistent entropy}, have proven valuable for extracting global, deformation-invariant features from data. Initially used for feature extraction, TDA provides stable and noise-robust descriptors based on the birth and death of topological features (e.g., connected components, loops) across scales, that can be used as input data for machine learning models across diverse tasks, such as immune networks characterization \cite{rucco2016characterisationpersistententropy} and time series comparison \cite{rucco2017newtopologicalentropyformeasuringsimilarities}. More recently, the field of \emph{topological deep learning} (TDL) has emerged, incorporating persistent homology into the training of neural networks via differentiable topological layers and loss functions \cite{TopologicalDeepLearning2024}. For instance, persistent homology and associated descriptors have been used to inject shape-aware constraints into neural networks, yielding improvements in tasks like anatomical image segmentation \cite{MALYUGINA2023TologicalLossFunctionforImageDenoisisng}, point cloud reconstruction \cite{hofer2017DeepLearningTopologicalSignatures}, shape classification \cite{Roell24aDifferentiableECTforshapeclassification}, and time series forecasting \cite{lin2024hybridizationpersistenthomologyneural}. Recent advances in differentiable topological optimization \cite{Mathieu2025DiffeomorphicInterpolationforEfficientPersistenceBasedTopologicalOptimization} have further enhanced the trainability and robustness of topological loss functions, making them viable components for neural network training pipelines. However, challenges such as gradient sparsity and computational overhead continue to hinder widespread adoption.

In this paper, we bridge geometry and topology to propose a \emph{novel, interpretable, and sustainable framework for function approximation under limited resources}. Our approach centers on the \emph{Barycentric Neural Network} ($\BNN$), a shallow, compact architecture that exactly represents continuous piecewise linear functions ($\CPLF$s) with guaranteed continuity across regions. Since any continuous function over a compact domain can be uniformly approximated by $\CPLF$s, the $\BNN$ naturally emerges as a flexible, interpretable, and efficient tool for function approximation. Crucially, the $\BNN$’s structure and parameters are fully determined by a set of trainable base points and their barycentric coordinates. Rather than optimizing internal weights, training reduces to optimizing the positions of these base points, turning the learning problem into a geometric one. The number of base points directly controls model capacity: fewer points yield lightweight models, while more enable higher fidelity, offering a natural trade-off between expressiveness and resource usage. To guide this geometric optimization, we introduce length-weighted persistent entropy ($\LWPE$), a new topological summary that extends classical persistent entropy by weighting each feature’s contribution by its persistence (lifetime). By emphasizing features that persist across multiple scales, this approach naturally prioritizes the most structurally important characteristics of the target function. This yields a stable, scale-invariant descriptor. We incorporate $\LWPE$ into a topologically informed loss function that preserves not only pointwise accuracy but also the global shape of the target function, something traditional losses like $\MSE$, $\RMSE$, $\MAE$, or LogCosh often neglect.

Our framework, $\BNN$ trained with $\LWPE$-based loss, is particularly effective in low-resource settings, such as when only a small number of base points or few training epochs are available. Experimental results on synthetic and real-world data demonstrate that our method achieves faster convergence and superior approximation quality compared to standard losses when limited resources are available, aligning it with the goals and principles of Green AI. This is especially valuable in domains where topological fidelity matters, such as preserving peaks and valleys in economic time series, motifs in biological signals, or global trends in physical simulations.

The key contributions of this work are as follows:

\begin{itemize}
    \item \textbf{Barycentric Neural Network}: A shallow and scalable architecture that exactly represents $\CPLF$s. Its structure and parameters are fixed, based on a trainable and predefined set of points and their barycentric coordinates. 
    \item \textbf{Geometric training via base points optimization}: The $\BNN$’s training is entirely based on optimizing the positions of its base points, yielding a geometrically and more efficient training process. 
    \item \textbf{Lenght-weighted persistent entropy}: A stable topological summary that weights features by persistence, enabling a more informative topological loss.
    \item \textbf{Experimental validation under resource constraints:} Comprehensive experiments showing that $\BNN$s trained with $\LWPE$ outperform classical losses (such as $\MSE$, $\RMSE$, $\MAE$ or LogCosh) in function approximation performance and convergence speed when base points and training epochs are limited.
\end{itemize}

By unifying geometric representation with topological regularization, our work contributes to the growing fields of topological deep learning and Green AI, while offering a practical, sustainable, and interpretable alternative to overparameterized models in function approximation tasks.

The remainder of this paper is organized as follows: Section \ref{sec:background} reviews key concepts involved in the paper, such as barycentric coordinates, persistent homology and persistent entropy. Section \ref{sec:persistentEntropy} introduces \emph{length-weighted persistent entropy}. Section \ref{sec:BNN} details how \emph{barycentric neural networks} are defined. In Section \ref{sec:experiments}, we present the experiments conducted, including an explanation of how our length-weighted persistent entropy-based loss function is defined, and how we use it for function approximation in combination with the barycentric neural network in low-resource scenarios, along with the results obtained. Finally, conclusions and future work are given in Section \ref{sec:conclusion}.

\section{Preliminaries}\label{sec:background}

This section introduces the main concepts and establishes some notations that are used throughout the paper. We begin by introducing $\CPLF$, followed by the key concepts of computational topology employed in this work, including simplicial complexes, barycentric coordinates, persistent homology, and persistent entropy. A general introduction to computational topology can be found, 
we refer the reader to \cite{edelsbrunner2022computationaltopology,Otter2017ComputationPersistentHomology}. 

\subsection*{\textbf{Continuous Piecewise Linear Functions}} 

Given $d\in\N$, a \textit{continuous piecewise linear function} ($\CPLF$) is a function $h:C \subseteq {\mathbb R}^d \to {\mathbb R}$, where the domain $C$ is the union of specific regions $|\sigma_0|,\dots,|\sigma_n|$ (called $d$-simplices, defined later). On each simplex \( |\sigma_k| \), the function is affine:
\[
h(p) = m_k \cdot p + c_k, \quad \text{for } p \in |\sigma_k|,
\]
with \( m_k \in \mathbb{R}^d \) and \( c_k \in \mathbb{R}^d \). Global continuity is enforced by requiring that adjacent linear pieces agree on shared faces: if \( p \in |\sigma_i| \cap |\sigma_j| \), then \( m_i \cdot p + c_i = m_j \cdot p + c_j \).

Finally, given any continuous function $f\colon \R^d\to\R$, a $\CPLF$ $h\colon C\subset \R^d\to\R$ approximates $f$ on $C$ with approximation error given by: 
 \[
 ||f-h||=\max\big\{|f(p)-( m_k \cdot p+c_k)|\colon p\in C \cap |\sigma_k| \mbox{ and } k\in\{0,1,\dots,n\}\big\}\,.
 \]

\begin{rem}\label{re:AB}
In our experiments (Section~\ref{sec:experiments}), we restrict attention to one-dimensional domains: \( h \colon [A,B] \subset \mathbb{R} \to \mathbb{R} \). Here, the domain is partitioned into subintervals \( [a_i, b_i] \) with \( A = a_0 < b_0 = a_1 < \cdots < b_{n-1} = a_n < b_n = B \), and continuity at breakpoints ensures \( m_{i-1} \cdot b_{i-1} + c_{i-1} = m_i \cdot a_i + c_i \). Any given continuous function $f:[A,B]\subset\R\to \R$ can be approximated as closely as desired by a $\CPLF$ $h:[A,B]\to \R$ by partitioning $[A,B]$ into a a sufficiently large finite number of segments.
\end{rem}

\subsection*{\textbf{Simplicial Complexes}}

Given $d\in\N$, a \emph{(finite, $d$-dimensional, embedded in $\mathbb{R}^d$) simplicial complex} $\K$ consists of:
\begin{itemize}
    \item A finite set of points $V\subset \R^d$ called vertices or 0-simplices,
    \item For $k\in\{1,2,\dots,d\}$, a finite set of $k$-simplices. Any $k$-simplex $\sigma$ of $\K$ is the convex hull of $k+1$ affinely independent vertices $\{v_0, \ldots, v_k\}\subset V$ (i.e., in general position). By abuse of notation, we write $|\sigma|$ to express the convex hull of the vertices of $\sigma$ and $\sigma=\{v_0,\dots,v_k\}$ to detail its vertices.
    \item Any $k$-simplex $\sigma\in\K$ has $k+1$ faces. Each face is the $(k+1)$-simplex obtained by removing one vertex of $\sigma$.
    \item If $\sigma \in \K$, all faces of $\sigma$ also belongs to $\mathcal{K}$.
    \item The non-empty intersection of two simplices of $\K$ is a face of both.
\end{itemize}

Any compact subset $\C$ of $\R^d$ can be approximated by a simplicial complex (see, for example, \cite[Chapter 1]{Munkres2000Topology}). The combinatorial structure of such complexes allows for efficient computation of topological information using discrete algorithms.

%%%%%%%%%%%%%%%%%%%%%%%%%%%%%%%%%%%%

\subsection*{\textbf{Barycentric Coordinates}}

Barycentric coordinates naturally connect geometry to topology through simplicial complexes. In any dimension $d$, barycentric coordinates $(t_0, t_1, \ldots, t_d)$  describe the location (position) of a point $p \in\R^d$ within the vertices of a $d$-simplex, ensuring that $t_i \geq 0$ for all $i\in\{0,1,\dots,d\}$ if and only if $p$ lies in $|\sigma|$. Besides, barycentric coordinates representation allows a continuous interpolation across the simplices of a simplicial complex. Formally, for any simplex $\sigma=\{v_0, v_1, \ldots, v_d\}$ with vertices $v_i\in\R^d$ in general position and any point $p\in |\sigma|\subset \R^d$, there exists unique real values $(t_0, t_1, \ldots, t_d)$ such that: 
\begin{equation*}
p = \sum_{i=0}^d t_i v_i,
\quad\mbox{with $t_i \geq 0$ and $\sum_{i=0}^d t_i = 1$.}
\end{equation*}
 
This representation allows a continuous interpolation across the simplices of a simplicial complex. Furthermore, unlike in \cite{PALUZOHIDALGO2020TwohiddenlayerFeedforwardNetworkareUniversalApproximators}, where the barycentric coordinates $(t_0,t_1,\dots,t_d)$ of a point $p\in\R^d$ with respect to a simplex $\sigma=\{v_0,v_1,\dots$, $v_d\}$ are computed using a matrix inverse, we instead employ the following equivalent formulation, which expresses the coordinates as determinants via Cramer’s rule:
 \[t_i=\frac{\lambda_i}{\sum_{j=0}^d\lambda_j},\qquad
 \lambda_i=\det\left(
 \begin{array}{ccccccc}
v_0&\cdots&v_{i-1}&p&v_{i+1}&\cdots&v_d\\
1&\cdots&1&1&1&\cdots&1
\end{array} \right),
\qquad 
 i=0,1,\dots,d.
 \]
This formulation gives explicit, differentiable, local, and interpretable expressions for barycentric coordinates. Moreover, it can be implemented as a neural network layer where each $\lambda_i$ serves as a linear score, all scores are computed independently, and the final normalization acts as a softmax operator without the exponential.

\begin{rem}\label{re:t}
The barycentric coordinates of a point $x\in\R$ with respect to an interval (i.e., a 1-simplex) $[a, b]\subset\R$ are $(1-t, t)$ with $t=\frac{x-a}{b-a}$. Then, $(1-t,t)\in[0,1]^2$ if and only if $x\in[a,b]$.
\end{rem}

\subsection*{\textbf{Lower-Star Filtration}}

Fixed $d\in\N$, let us consider a simplicial complex $\K$ with real values assigned to its vertices $V\subset \R^d$, denoted by $g \colon \R^d \to \mathbb{R}$. Assuming $V$ is composed by vertices $\{u_0, u_1, \dots,u_m\}$ with distinct values $g(u_0) < g(u_1) < \ldots < g(u_m)$, the \emph{lower-star} of a vertex $u\in V$ is the subcomplex that consists of all the simplices $\sigma$ of $\K$ such that $u$ is the minimum vertex of $\sigma$, that is $u\in\sigma$ and $g(u)<g(v)$ for any $v\in\sigma$. The \emph{lower-star filtration of $\K$} is the increasing sequence of simplicial complexes $\K(1)\subset \K(2)\subset\cdots\subset \K(n)=\K$ where $\K(i)$ is the lower-star of vertex $u_i$. See \cite[Chapter VI]{edelsbrunner2022computationaltopology}. As we will see later, this filtration will be used to compute persistent homology, and consequently the topological loss based on $\LWPE$ used in the experiments of this paper.

\subsection*{\textbf{Persistent Homology, Persistence Diagrams and Barcodes}}

Persistent homology is a tool for studying data geometry and connectivity across scales. Using this tool and given a \emph{filtered simplicial complex} (e.g., lower-star filtration), one can effectively compute $q$-dimensional topological features at different scales, tracking features as they appear (birth) and disappear (death). Persistence diagrams and barcodes are visual representations used in the study of persistent homology and represent the birth and death (end) of topological features across scales. A $q$-dimensional persistence diagram plots points $(b, e)$ corresponding to the lifespans of $q$-dimensional topological features, while a barcode represents them as a {\it bar} starting at $b$ and ending at $e$. Specifically, a $q$-dimensional persistence diagram $D = \{(b_i, e_i) \mid i \in I\}$ satisfies that $b_i \leq e_i$ for all $i \in I$, and $I$ is the index set that identifies the pairs $(b_i,e_i)$ in $D$. Each pair $(b_i,e_i)$ corresponds to a $q$-dimensional topological feature that {\it appears} at time $b_i$ (birth) and {\it disappears} at time $e_i$ (death, i.e., end) as the filtration progresses. Both persistence diagrams and barcodes facilitate the understanding of feature persistence in data. Fig. \ref{fig:entropyCalculation} shows a persistence diagram (center) alongside its corresponding barcode (right), illustrating the persistence of topological features in the data.

The number of bars of the barcode associated with the lower-star filtration of a simplicial complex $\K$ is less than or equal to half of the number of vertices of the simplicial complex $\K$ plus 1. Given the lower-star filtration on a simplicial complex $\K$ with vertex set $V$, local critical points are always located at the vertices of $V$ and can be effectively computed from the {\it lower link} of each vertex $v$ of $V$, which consists of the simplices of $\K$ in the closed lower-star that do not belong to the lower-star of $v$. We call $v$ a local critical vertex of index $q$ if its lower link has the reduced homology of the $(q -1)$-sphere. Then, $q$-dimensional persistence diagrams pair critical vertices of index $q$ (when topological features are born) with critical vertices of index $q+1$ (when topological features die). A detailed description can be found in~\cite[Chapter VI]{edelsbrunner2008persistenthomology}.

\begin{rem}
In the case of 0-dimensional persistent homology (the one considered in the experiments of this paper), the number of topological features (number of points in persistent diagram, and number of bars in persistent barcode) is tight: with $2n-1$ vertices, one can obtain a $0$-dimensional barcode with $n$ bars since local minima (vertices where connected components are born) are paired with local maxima (vertices where connected components die) through the 0-dimensional persistence diagrams and barcodes. By convention, we truncate the death of the oldest connected component to $\max f$. This way, when we use the lower-star filtration, the longest bar is always $[\min f,\max f)$. Figure~\ref{fig:entropyCalculation} illustrates this with a simplicial complex composed of vertices (points) and edges (union between points) whose function $f$ is derived from the sine function, producing four local minima paired with maxima, represented by four bars. This way, we can obtain the same barcode, just with seven vertices (being four local minima and three local maxima).
\end{rem}

\subsection*{\textbf{Persistent entropy}}

Persistent entropy ($\PE$) \cite{chintakunta2015entropybarcode} {\it measures} the complexity of a topological space based on its persistence diagram $D = \{(b_i, e_i) \mid i \in I\}$. It is defined as :
\begin{equation}
\PE = - \sum_{i \in I} p_i \ln{p_i},\quad \mbox{ where $p_i = \frac{\ell_i}{L}$, being $\ell_i = e_i - b_i$ and $L = \sum_{i \in I} \ell_i$.}
\end{equation} 

\begin{figure}[ht!]
     \centering
     \includegraphics[width=0.95\textwidth
     ]{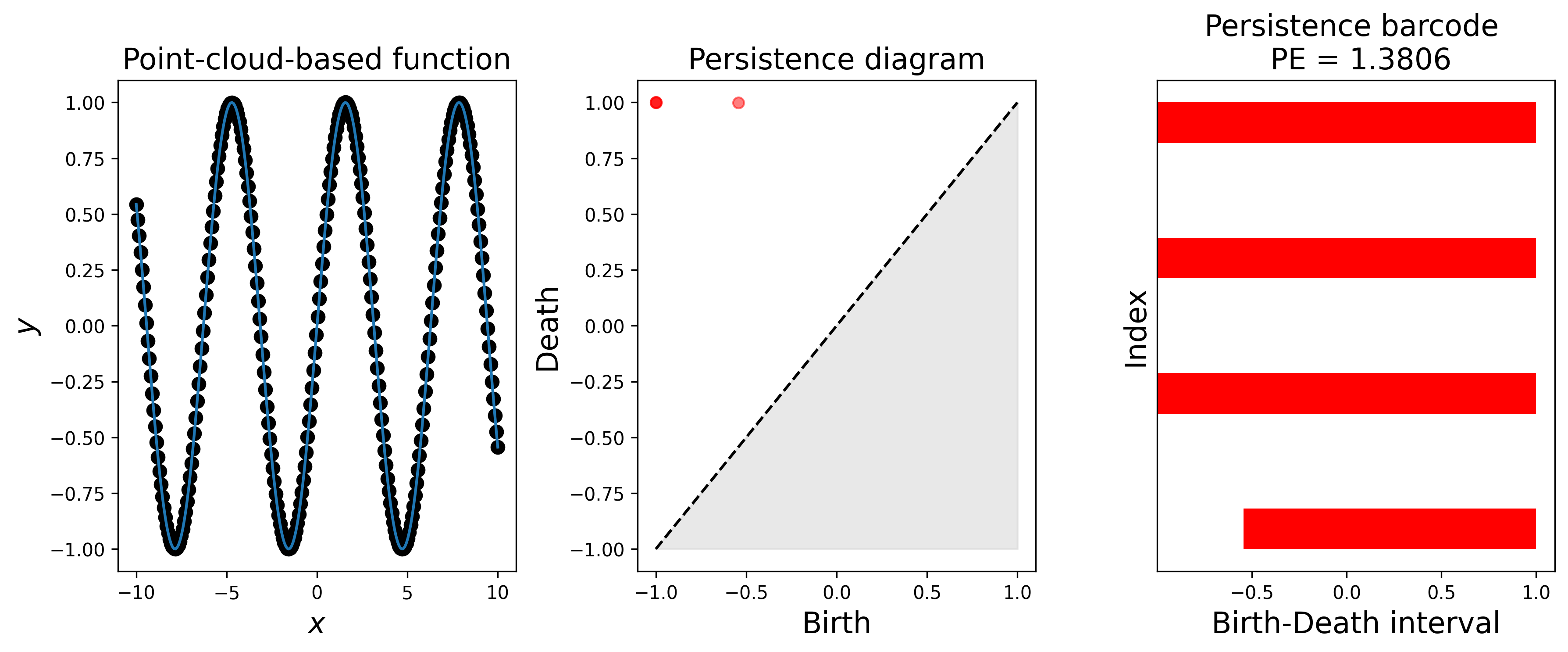}
    \caption{The figure shows a simplicial complex with 250 vertices (point-cloud-based function with 250 points) representing the sine function (left), its persistence diagram (middle), barcode, and persistent entropy (right), all from the lower-star filtration.} \label{fig:entropyCalculation}
\end{figure}

Persistent entropy reflects the uniformity of bar lengths in a persistence barcode. Its maximum is attained when all bars have equal length, in which case it is given by $\PE = \ln(\#I)$, where $\#I$ denotes the cardinality of $I$ (i.e., the number of points in the persistence diagram $D$). Conversely, its minimum value is $\PE = 0$, which occurs when the barcode contains a single long bar and the remaining bars have negligible length. Moreover, the greater the disparity among bar lengths, the closer the value of persistent entropy is to its minimum.

The stability results on persistence diagrams~\cite{cohen-steiner2007stability} and persistent entropy~\cite{atienza2020stabilitypersistententropy,rucco2017newtopologicalentropyformeasuringsimilarities} provide a robust foundation for quantifying the stability of topological features under small perturbations. This property is particularly valuable in applications where small variations in data or model outputs, such as those encountered in numerical approximations or machine precision errors, might otherwise introduce significant uncertainties in analytical results. As an illustration, the persistent entropy computed in the example shown in Fig.~\ref{fig:entropyCalculation} is 1.3806.

\section{\textbf{Lenght-Weighted Persistent Entropy}}\label{sec:persistentEntropy}

Persistent entropy ($\PE$) computed via lower-star filtration is suitable for comparing discrete piecewise linear functions and analyze time series \cite{rucco2016characterisationpersistententropy}. However, PE suffers from a fundamental limitation: it is invariant under uniform rescaling of the function values. This occurs because $\PE$ depends only on the \emph{relative proportions} of bar lengths in the persistence diagram, not their absolute magnitudes. Consequently, two functions with markedly different scales or resolutions can yield nearly identical $\PE$ values, even when their geometric or topological fidelity differs substantially. 

To overcome this limitation,we introduce \emph{length-weighted persistent entropy} ($\LWPE$), a variant that preserves the entropy structure while incorporating absolute persistence information. Formally, $\LWPE$ is defined as: 
\begin{equation} 
\LWPE = - \sum_{i \in I} \ell_i \ln{p_i} ,\quad \mbox{ where $p_i = \frac{\ell_i}{L}$, $\ell_i = e_i - b_i$ and $L = \sum_{i \in I} \ell_i$.} 
\end{equation} 

Unlike $\PE$, which is scale-invariant, $\LWPE$ explicitly weights each term by the absolute persistence \( \ell_i \). This ensures that longer-lived (i.e., more topologically significant) features contribute more substantially to the overall entropy, making $\LWPE$ sensitive to both the \emph{distribution} and the \emph{scale} of topological features.

\begin{figure}[ht!]
     \centering \includegraphics[width=0.95\textwidth]{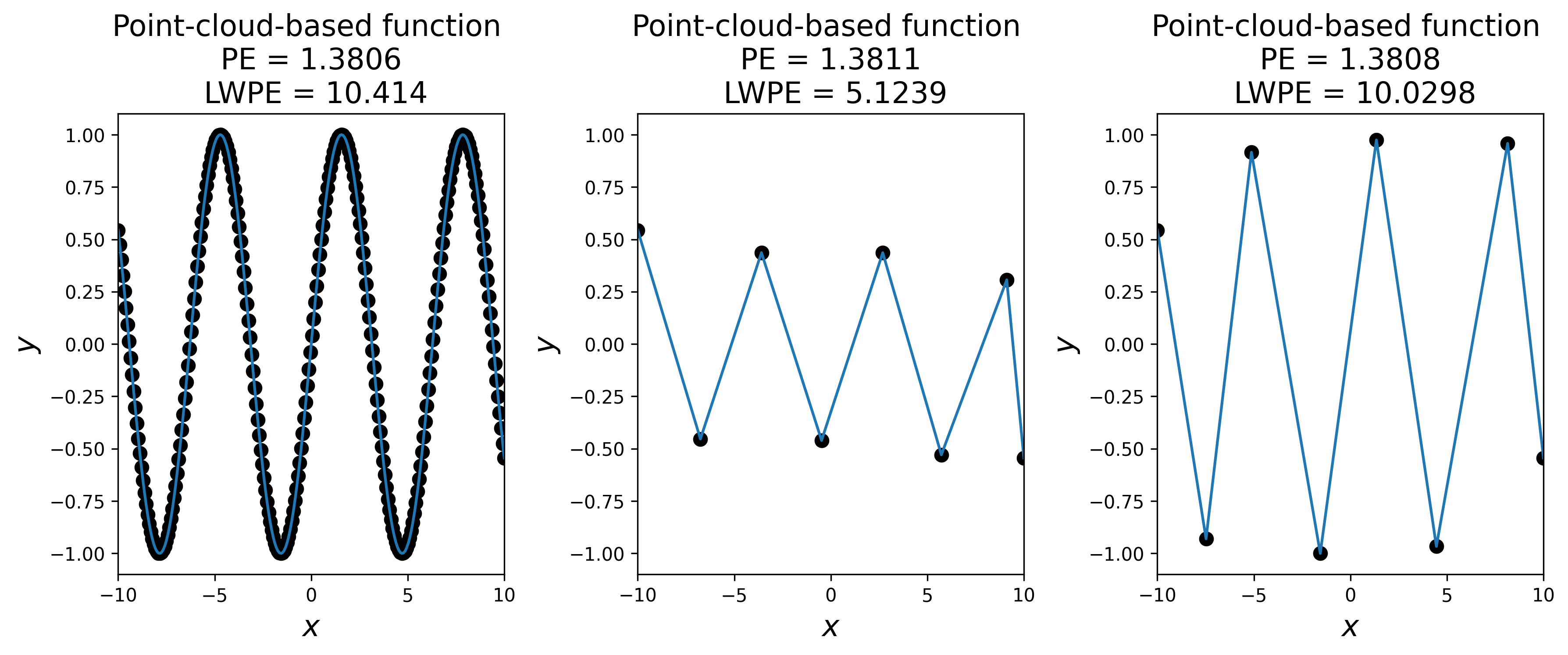}
    \caption{Comparison of persistent entropy ($\PE$) and length-weighted persistent entropy ($\LWPE$) on three subsampled versions of a sine function. Despite similar $\PE$ values across resolutions, $\LWPE$ distinguishes coarse approximations (middle) from faithful ones (right), reflecting differences in absolute topological scale.} \label{fig:limitationPE}
\end{figure}

Figure~\ref{fig:limitationPE} illustrates this difference. Three subsampled versions of the same sine function are shown. Although the relative shape of the signal is preserved, the middle plot uses so few points that the resulting piecewise linear approximation poorly captures the true peak structure. Standard $\PE$ remains nearly unchanged across all three cases because it only reflects the normalized distribution of bar lengths. In contrast, $\LWPE$ drops significantly in the middle case, correctly indicating a loss of topological fidelity due to reduced resolution. Only when both the relative proportions and the absolute scale of features are preserved (right plot) does $\LWPE$ remain close to the reference value. 

As shown later in Section~\ref{sec:experiments}, using a topological loss function based on $\LWPE$ leads to substantially improved approximation quality, particularly in low-resource settings, by guiding the optimization of base points toward solutions that preserve not only the shape but also the scale of topologically relevant structures.

\section{Barycentric Neural Networks}\label{sec:BNN}

A \emph{neural network} is typically viewed as a parametrized mathematical function $N_\theta: \mathbb{R}^d \to \mathbb{R}^m$, where $\theta$ encompasses weights, biases, activation functions, and architectural choices such as depth and layer types \cite{gurney2018introductionNNs}. In this paper, we focus on \emph{multi-layer feed-forward networks}, whose hidden layers compute transformations of the form $y = h(Wx + b)$, with input $x \in \mathbb{R}^d$, weight matrix $W$, bias vector $b$, and nonlinear activation $h$ (e.g., ReLU).

We now introduce the \emph{Barycentric Neural Network} ($\BNN$), a shallow architecture that leverages barycentric coordinates to represent continuous piecewise linear functions exactly. The key idea is to replace trainable weights with geometrically meaningful \emph{base points}, turning the learning problem into an optimization over point positions.

\begin{defn}\label{def:bnnsigma}
Given a function $g:\R^d\to \R$ and a $d$-simplex $\sigma=\{v_0,\dots,v_d\}$ with vertices $v_i\in\R^d$, the {\it barycentric neural network $\BNN_{\sigma}$} associated with $\sigma$ is defined as: 
\[\BNN_{\sigma}(p) = \sum_{i=0}^d\ReLU\Big(1-\ReLU(1-t_i)-\sum_{j=0}^d\big(\step^*(-t_j)+\step^*(t_j-1)\big)\Big)\cdot g(v_i)\,,\]
where $p\in\R^d$, $t_i=\frac{\lambda_i}{\sum_{j=0}^d\lambda_j}$, $\lambda_i=\det\left(
 \begin{array}{ccccccc}
v_0&\cdots&v_{i-1}&p&v_{i+1}&\cdots&v_d\\
1&\cdots&1&1&1&\cdots&1
\end{array} \right)$, $i=0,1,\dots,d$, ReLU corresponds to the activation function defined by $\ReLU(t) = \max\{0,t\}$, for $t\in\R$, and $\step^*$ is the activation function defined as $\step^*(t)=1-\step(-t)=\left\{\begin{array}{cl}1 & \mbox{ if $t>0$}\\0&\mbox{ if $t\leq 0$}\end{array}\right.$ where $\step(t) = 1$ if $t\geq 0$ and 0 otherwise.
\end{defn}

\begin{lem}\label{lem:1}
Let $\sigma=(v_0,v_1,\dots,v_d)$ be a $d$-simplex with vertices $v_i\in\R^d$. Let $p\in\R^d$ with barycentric coordinates $(t_0,t_1,\dots,t_d)$ with respect to $\sigma$. We have:
\begin{itemize}
\item[\textnormal{(a)}]
If $p\in |\sigma|$ then $\BNN_{\sigma}(p)=\sum_{i=0}^d t_i\cdot g(v_i)$.
\item[\textnormal{(b)}]
If $p\not\in|\sigma|$ then $\BNN_{\sigma}(p)=0$.
\end{itemize}
\end{lem}

\begin{proof}
If $p\in|\sigma|$  then $t_i\in[0,1]$ for all $i\in\{0,1,\dots,d\}$ by definition of barycentric coordinates. Then $\step^*(-t_j)=0= \step^*(t_j-1)$ for all $j\in\{0,1,\dots,d\}$ and therefore,
 \[\ReLU\Big(1-\ReLU(1-t_i)-\sum_{j=0}^d\big(\step^*(-t_j)-\step^*(t_j-1)\big)\Big) = \ReLU\Big(1-\ReLU(1-t_i)\Big)=t_i.\]
We then have that $\BNN_{\sigma}(p) = \sum_{i=0}^d t_i\cdot g(v_i)$.
\\
\\
\noindent
If $p\not\in |\sigma|$ then, by the definition, the barycentric coordinates $(t_0,t_1,\dots,t_d)$ of $p$ with respect to $\sigma$ satisfy that there is a non-emtpy set $J\subset \{0,1,\dots,d\}$ such that for any $j\in J$, $t_j<0$ or $t_j>1$. In that case, $\step^*(-t_j)=1$ and $\step^*(1-t_j)=0$ when $t_j<0$, or $\step^*(-t_j)=0$ and $\step^*(t_j-1)=1$ when $t_j>1$. Therefore, $\step^*(-t_j)+\step^*(t_j-1)=1$. Then,
\[
\sum_{j=0}^d\big(\step^*(-t_j)+\step^*(t_j-1)\big)\geq 1\]
and therefore, 
\[\ReLU\Big(1-\ReLU(1-t_i)-
\sum_{j=0}^d\big(\step^*(-t_j)+\step^*(t_j-1)\big)\Big)=0\]
for all $i\in\{0,1,\dots,d\}$, concluding that $\BNN_{\sigma}(p)=0$.  
\end{proof}

\begin{rem}\label{re:bnn}
Consider the $1D$ case of a function $g:\R\to \R$ and a 1-simplex $\sigma=\{a,b\}$ with $a<b\in \R$. We have that the barycentric coordinates of a point $x\in\R$ with respect to $\sigma$ are $(1-t,t)$, being $t=\frac{x-a}{b-a}$. Then,
\begin{align*}
    \BNN_{\sigma}(x)
    %\big((1-t,t)\big) 
    &= \ReLU\big(1 - \ReLU(t) - (\step^*(t-1) + \step^*(-t)\ + \step^*(-t) + \step^*(t-1)) \big) \cdot g(a)
    \\
    &+ \ReLU\big(1-\ReLU(1-t) - (\step^*(t-1) + \step^*(-t)+ \step^*(-t) + \step^*(t-1)) \big) \cdot g(b)
    \\ 
 &= \ReLU\Big(1 - \ReLU(t) - 2 \cdot (\step^*(t-1) + \step^*(-t)\big) \Big) \cdot g(a)
    \\
    &+ \ReLU\big(1-\ReLU(1-t) - 2\cdot( \step^*(t-1) + \step^*(-t))\big) \cdot g(b)\,
    \\
    &= \ReLU\big(1 - \ReLU(t) - 2\step^*(t-1) -2 \step^*(-t)\big) \cdot g(a) 
    \\
    &+ \ReLU\big(1 - \ReLU(1-t) - 2\step^*(t-1) -2 \step^*(-t)\big) \cdot g(b).
\end{align*}
Therefore, $\BNN_{\sigma}(x)=0$ if $t<0$ or $t> 1$, that is, if $x$ is outside the closed interval $[a,b]$, ensuring locality.
\end{rem}

\begin{figure}[ht!]
\centering
\includegraphics[width=0.95\textwidth]{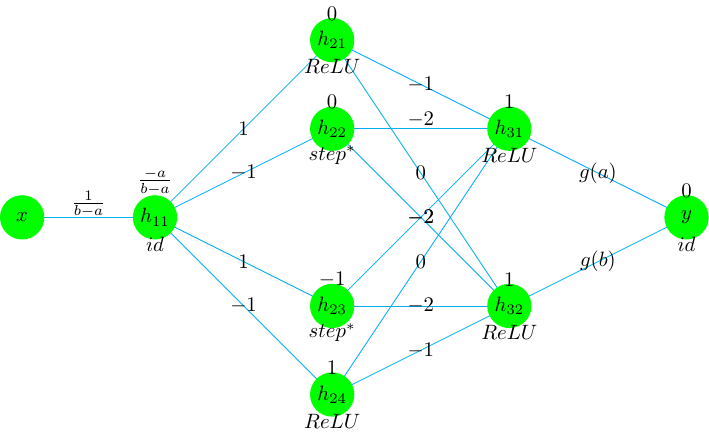}
\caption{Structure and parameters of $\BNN_{\sigma}$ in the case of $\sigma=\{a,b\}$, with $a<b\in\R$, $x\in\R$. Each hidden neuron is denoted as $h_{ij}$, where $i$ indicates the hidden layer and $j$ the neuron index within that layer (e.g., $h_{22}$ is the second neuron of the second hidden layer). The numbers written above the neurons correspond to biases, while the labels below indicate the activation functions. The numbers on the edges represent weights. The first hidden layer only aims to compute the barycentric coordinate $t$ within the simplex $\sigma$ , while the subsequent layers determine the contribution of the point within that simplex.}
\label{fig:bnnillustration}
\end{figure}

As we can see in Fig. \ref{fig:bnnillustration}, in the case of $\sigma=\{a,b\}$ with $a<b\in \R$, $\BNN_{\sigma}$ corresponds to a multi-layer feedforward network with a single-neuron input and output layer (identity activation, referred to as $id$). It has three hidden layers with one, four and two neurons, respectively, where neurons in the same layer may have different activation functions.

\begin{defn}\label{def:bnn}
Let $C = \bigcup_{k=0}^n |\sigma_k|$ where \(\sigma_k = \{v_0^k, v_1^k, \ldots, v_d^k\}\) are the \(d\)-simplices of a \(d\)-dimensional simplicial complex \(K\), and let \(g : \R^d \to \R\). Then, for $p\in\R^d$
we define the global $\BNN$ as:

\[
\mathrm{BNN}(p) = \frac{1}{\#\{k : \mathrm{BNN}_{\sigma_k}(p)> 0\}}\cdot \sum_{k=0}^{n} \mathrm{BNN}_{\sigma_k}(p)
,
\]
where \(\#\{k : \mathrm{BNN}_{\sigma_k}(p)> 0\}\) counts the number of local networks \(\mathrm{BNN}_{\sigma_k}\) that are nonzero at the barycentric coordinates $(t^k_0,t^k_1,\dots,t^k_d)$ of $p$ with respect to $\sigma^k$. 
\end{defn}

In other words, the value of \(\mathrm{BNN}\) at a point \(p\) is the average of the values of the local networks \(\mathrm{BNN}_{\sigma_k}\) for which \(p\) lies in \(|\sigma_k|\), which is precisely detected by \(\mathrm{BNN}_{\sigma_k}(p)> 0\). This normalization properly accounts for overlapping simplices: points at the intersection of multiple simplices (e.g., shared vertices or edges) yield multiple nonzero local evaluations, and the $\BNN$ formula averages them to ensure continuity and consistency.

\begin{example}
    Suppose $g:[A,B]\to\R$ where  $\sigma_1=\{a_1 = A,a_2=b_1\}$ and $\sigma_2=\{b_1=a_2,b_2=B\}$ are two adjacent 1-simplices, we have that the barycentric coordinates of $b_1=a_2$ with respect to $\sigma_1$ are $(0,1)$ and with respect to $\sigma_2$ are $(1,0)$. Then
\begin{align*}
    \BNN_{\sigma_1}(a_2)
    =& 
 \ReLU\big(1-\ReLU(1)-2\cdot\step^*(0)-2\cdot \step^*(-1)\big)
\cdot g(a_1)\\
&+
\ReLU\big(
1-\ReLU(0)-2\cdot\step^*(-1)-2\cdot \step^*(0)\big) \cdot g(a_2)=g(a_2)\,
\end{align*}
\begin{align*}
    \BNN_{\sigma_2}(b_1)
    =&
 \ReLU\big(1-\ReLU(0)-2\cdot\step^*(-1)-2\cdot \step^*(0)\big)
\cdot g(b_1)\\
&+
%\big(
\ReLU\big(1-\ReLU(1)-2\cdot\step^*(0)-2\cdot \step^*(-1)\big) \cdot g(b_2)=g(b_1).\,
\end{align*}
Then, $\BNN_{\sigma_1}(a_2) +\BNN_{\sigma_2}(b_1) = g(a_2) + g(b_1) = 2\cdot g(b_1)=2\cdot g(a_2)$ (since $b_1=a_2)$. This illustrates the redundancy at joining points and motivates the normalization in the global BNN.
\end{example}

We present the following result, ensuring consistency, continuity and locality of BNNs.

\begin{thm}\label{mainresult}
Let $C = \bigcup_{k=0}^n |\sigma_k|$ where \(\sigma_k = \{v_0^k, v_1^k, \ldots, v_d^k\}\) are the \(d\)-simplices of a \(d\)-dimensional simplicial complex \(K\), and let \(g : \R^d \to \R\). We have: 
\begin{enumerate}
\item[\textnormal{(1)}] $\BNN(v^k_{\ell})=g(v^k_{\ell})$ for all $\ell\in\{0,1,\dots,d\}$ and $k\in\{0,1,\dots,n\}$.
\item[\textnormal{(2)}] If $p\in C$ then
\[\BNN(p)=\BNN_{\sigma^k}(p)=\sum_{i=0}^d t^k_i \cdot g(v^k_i). \] where $(t_0^k,t_1^k,\dots,t_d^k)$ are the barycentric coordinates of $p$ with respect to $\sigma_k$ for any $k$ such that $p\in|\sigma_k|$.
\item[\textnormal{(3)}] If $p\not\in C$ then
\(\BNN(p)=0\).
\item[\textnormal{(4)}] The function $\BNN:C\to\R$ is a CPLF.
\end{enumerate}
\end{thm}

\begin{proof}
First, recall the definition of $\BNN_{\sigma}$ for a give point $p\in\R^d$ with barycentric coordinates $(t^k_0,t^k_1,\dots,t^k_d)$ with respect to a simplex $\sigma^k$:
\[\BNN_{\sigma_k}(p)=
\sum_{i=0}^d\ReLU\Big(1-\ReLU(1-t^k_i)-\sum_{j=0}^d\big(\step^*(-t^k_j)+\step^*(t^k_j-1)\big)\Big)\cdot g(v^k_i).\]
Now, let us prove the statements. By definition, the barycentric coordinates of $v^k_{\ell}$ with respect to $\sigma_k$ are $(0,\stackrel{\ell}{\dots},0,1,0 \stackrel{d-\ell}{\dots},0)$ where the value $1$ is at position $\ell$.Since $\step^*(0)=step^*(-1)=0$ and $1-\ReLU(1-0)=0$ then
\[\BNN_{\sigma_k}(p)= \ReLU\Big(1-\ReLU(1-1)\Big)\cdot g(v^k_{\ell})=g(v^k_{\ell}).\]
Similarly, if $v^k_{\ell}=v^{k'}_{\ell'}$ for some $k'\in\{0,1,\dots,n\}$ and $\ell'\in\{0,1,\dots,d\}$ then 
\[\BNN_{\sigma_{k'}}(p)=
g(v^{k'}_{\ell'})=g(v^{k}_{\ell}).\]
We can then conclude that $\BNN(v^k_{\ell})=g(v^{k}_{\ell}).$
\\
\\
\noindent If $p\in|\sigma_k|$ then by Lemma~\ref{lem:1} we have that
\[\BNN_{\sigma_k}(p)=
\sum_{i=0}^d t^k_i\cdot g(v^k_i).\]
If $p\in|\sigma^i|\cap |\sigma^j|$ for some $i,j\in\{0,1,\dots,n\}$, let $(t^i_0,t^i_1,\dots,t^i_d)$ and $(t^j_0,t^j_1,\dots,t^j_d)$ be the barycentric coordinates of $p$ with respect to $\sigma_i$ and $\sigma_j$ respectively. By the definition of simplicial complexes, there is an $\ell$-simplex $\sigma=\{v_0,v_1,\dots,v_{\ell}\}\subset \sigma_i,\sigma_j$ with  $\ell<d$ such that $p\in |\sigma|$ and satisfying that there are sets of indices $I,J\subset\{0,1,\dots,d\}$ such that $v_r=v^i_{r_i}=v^j_{r_j}$ and $t_r=t^i_{r_i}=t^j_{r_j}$ for some $r_i\in I$ and $r_j\in J$ and for all $r\in\{0,1,\dots,\ell\}$, being $(t_0,t_1,\dots,t_{\ell})$ the barycentric coordinates of $p$ with respect to $\sigma$. Putting all together, 
\[\BNN_{\sigma^i}(p)=\sum_{r_i\in I}t^i_{r_i}\cdot g(v^i_{r_i})=
\sum_{r_j\in I}t^j_{r_j}\cdot g(v^j_{r_j})=
\BNN_{\sigma^j}(p)
.\]
We then conclude that, for any $k\in\{0,1,\dots, n\}$ with $p\in|\sigma_k|$, $\BNN(p)=\sum_{i=0}^d t^k_i\cdot g(v^k_i)$, where $(t^k_0,t^k_1,\dots,t^k_d)$ are the barycentric coordinates of $p$ with respect to $\sigma_k$.
\\
\\
\noindent
If $p\not\in C$ then, 
$p\not\in |\sigma_k|$ for any given $k\in\{0,1,\dots,n\}$. Therefore, by Lemma~\ref{lem:1}, 
$\BNN_{\sigma^k}(p)=0$ and so $\BNN(p)=0$.
\\
\\
\noindent
Let $p\in C$. Then $p\in |\sigma^k|$ for some $k\in\{0,1,\dots,n\}$. Let $(t^k_0,t^k_1,\dots,t^k_d)$ be the barycentric coordinates of $p$ with respect to $\sigma_k$,  by Lemma~\ref{lem:1} and by the definition of barycentric coordinates, we have:
\begin{align*}
\BNN(p)=\sum_{i=0}^dt^k_i\cdot g(v^k_i)
=&(\begin{array}{cccc}
g(v^k_0)&g(v^k_1)&\cdots&g(v^k_d)
\end{array})\cdot
(\begin{array}{cccc}
t_0&t_1&\cdots&t_d
\end{array})^{\scriptscriptstyle T}
\\
=&(\begin{array}{cccc}
g(v^k_0)&g(v^k_1)&\cdots&g(v^k_d)
\end{array})\cdot\left(\begin{array}{cccc}
v_0&v_1&\cdots&v_d\\
1&1&\cdots&1
\end{array}\right)^{-1}\cdot 
\left(\begin{array}{c}
p\\
1
\end{array}\right)
\\
=&(\begin{array}{ccccc}
m^k_0&m^k_1&\cdots&m^k_{d-1}&c_k
\end{array})\cdot
\left(\begin{array}{c}
p\\
1
\end{array}\right)= m_k\cdot p+c_k
\end{align*}
where $m_k\in\R^d$ and $c_k\in\R$, concluding the proof.
\end{proof}

A key constraint in this setup is the following:
\begin{rem}\label{re:constraint}
Given a fixed number $n$, a BNN with $n$ base points can only approximate functions $f:[A,B]\to\R$ (sampled by a point cloud $X$) whose persistence barcode contains at most $n/2$ bars, or, in the noisy case, the $n/2$ most significant bars.
\end{rem}

Still, BNNs are universal approximators.

\begin{rem}\label{re:BNN}
Any $\CPLF$ $g:[A,B]\to\R$ can be represented by a $\BNN$, with the key property that the $\BNN$ does not require an analytic definition. Instead, to specify the $\CPLF$ or the corresponding $\BNN$, it is sufficient to provide a finite set of transition (base) points $\{a_i\}\subset\R$, where the $\CPLF$ shifts between linear segments, together with the $\CPLF$ value at these points. This allows the $\BNN$ to capture the contribution of each segment relative to the input, thereby generating the correct output. For consistency with the terminology introduced in the Introduction (Section \ref{sec:intro}), in the following we shall refer to these transition points as \emph{base points}.
\end{rem}

\section{Testing BNNs and Length-Weighted Persistent Entropy-Based Loss in the Context of Resource-Efficient AI}\label{sec:experiments}

In this section, we test the capabilities of our proposed framework ($\BNN$ trained with $\LWPE$-based loss) in the context of \emph{Resource-Efficient AI}, assuming that we have limited resources for network design and training, such as a small number of base points and few training epochs. The goal is to approximate an objective function $f\colon [A,B]\to\R$, which is known on a finite set $S$ of points, where $\{A,B\} \subset S \subset [A,B]$, using just a limited number of base points to define the $\BNN$ that will approximate $f$. 

In our approach, the optimization is entirely carried out on these base points. Once their positions are fixed, both the network’s connectivity and its associated parameters are uniquely determined. Each base point acts as a generator in the continuous input domain, not constrained to coincide with the discrete set $S$ where $f$ is known. To guide this optimization, we propose using a topological loss function based on the \emph{length-weighted persistent entropy} ($\LWPE$). We compared it against classical losses, like $\MSE$, $\RMSE$, $\MAE$, or LogCosh, when used to optimize the same set of base points. 

That is to say, the training follows a standard iterative procedure organized in epochs, but the variables being optimized are the positions of the base points. At epoch zero, an initial configuration of base points is generated, and the corresponding $\BNN$ is automatically constructed. Using this configuration, the network produces an initial approximation of the target function at the sampled points $S$. The error between the predicted and true values is then evaluated, either through classical losses such as $\MSE$, $\RMSE$, $\MAE$, or LogCosh, or through our proposed topological loss based on the $\LWPE$. For the topological loss, persistent homology is computed using the lower star filtration for both the target function and the $\BNN$ approximation. The corresponding persistence barcodes are then used to calculate their length-weighted persistent entropies. The difference between these two entropy values quantifies how well the approximation preserves the topological features of the target function. Once the loss is computed, the framework estimates how small displacements of the base points affect the overall error, updating their positions accordingly.

In summary, although no explicit weights are directly optimized, the movement of the base points within the continuous domain implicitly determines the configuration of the $\BNN$ parameters. This makes the training conceptually equivalent to that of traditional neural networks, but constrained to a compact and geometrically structured search space.

In summary, although no explicit weights are ever optimized, the movement of the base points within the continuous input domain implicitly determines the configuration of the $\BNN$ parameters. This makes the training process conceptually equivalent to the traditional optimization of neural networks, but confined to a compact and geometrically structured search space.

We used the TensorFlow\footnote{The TensorFlow documentation is available at \url{https://www.tensorflow.org/}.} and Gudhi\footnote{The Gudhi documentation is available at \url{https://gudhi.inria.fr/}.} libraries to implement our proposed $\BNN$ and our proposed topological loss, as well as to conduct our experiments.

First, we begin by noticing that a $\BNN$, constructed using a set of equidistant base points, can effectively approximate any continuous nonlinear function $f:[A,B]\subset \mathbb{R} \to \mathbb{R}$, such as $f(x) = \sin(x)$. This relies on Remark~\ref{re:BNN}, which states that any $\CPLF$ can be represented by a $\BNN$, and Remark~\ref{re:AB} that states that a CPLF, and therefore a $\BNN$, acts as a universal approximator for continuous functions over compact intervals. 

In more realistic scenarios, the resources available for network design and training are often limited. Instead, we consider a function $f:[A,B]\subset \R \to \R$ that is represented by a point cloud, defined as a set $X=\{(x_i,y_i)\}_{i\in I}$, where each pair $(x_i,y_i)$ is a sample point, that is, $y_i=f(x_i)$. In general, we say that a set $X$ is a {\it point-cloud-based function} if $x_i = x_j$ implies that $y_i=y_j$ for all $i,j\in I$. We aim to approximate $X$ using a $\BNN$ built from a small set $S$ of base points that includes the endpoints $A$ and $B$. 

\begin{figure}[ht!]
    \centering
    \begin{subfigure}[b]{0.45\textwidth}
    \centering
    \includegraphics[width=\textwidth]{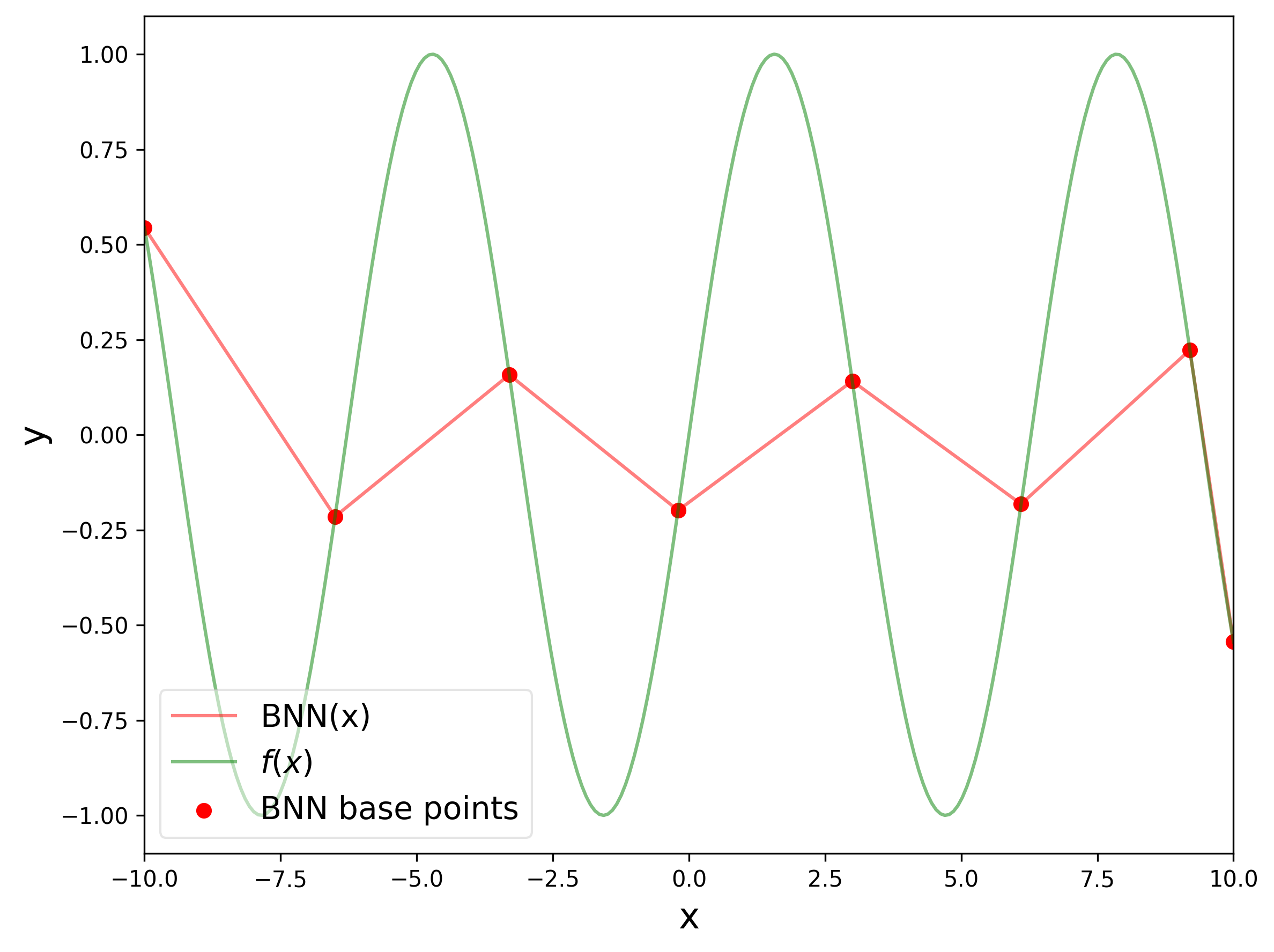}  
    \caption{}
    \label{fig:comparingPEsInitial}
    \end{subfigure}
    \begin{subfigure}[b]{0.45\textwidth}
    \centering
    \includegraphics[width=\textwidth]{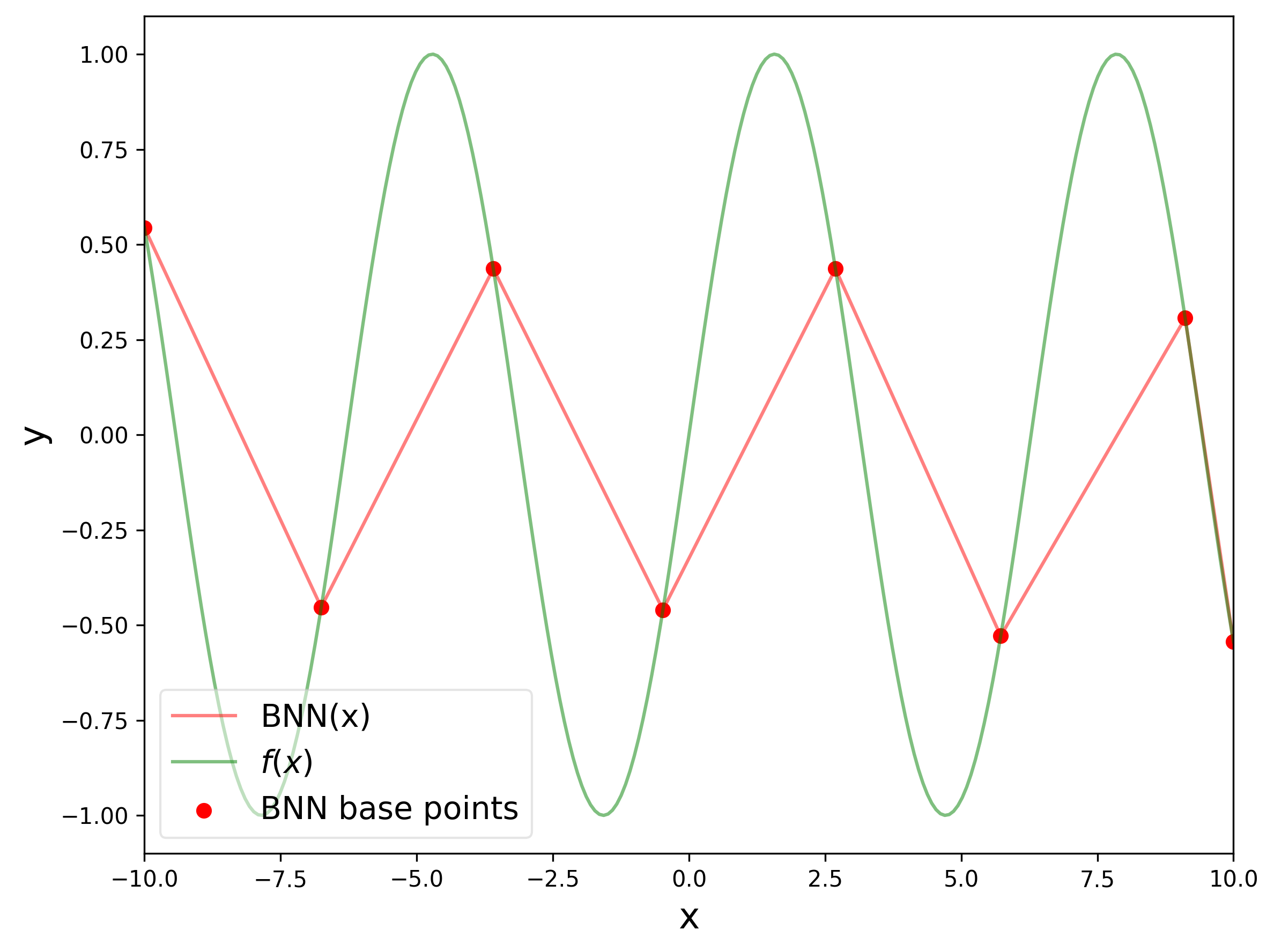}
    \caption{}
    \label{fig:comparingPEsAproxPE}
    \end{subfigure}
    \begin{subfigure}[b]{0.45\textwidth}
    \centering
    \includegraphics[width=\textwidth]{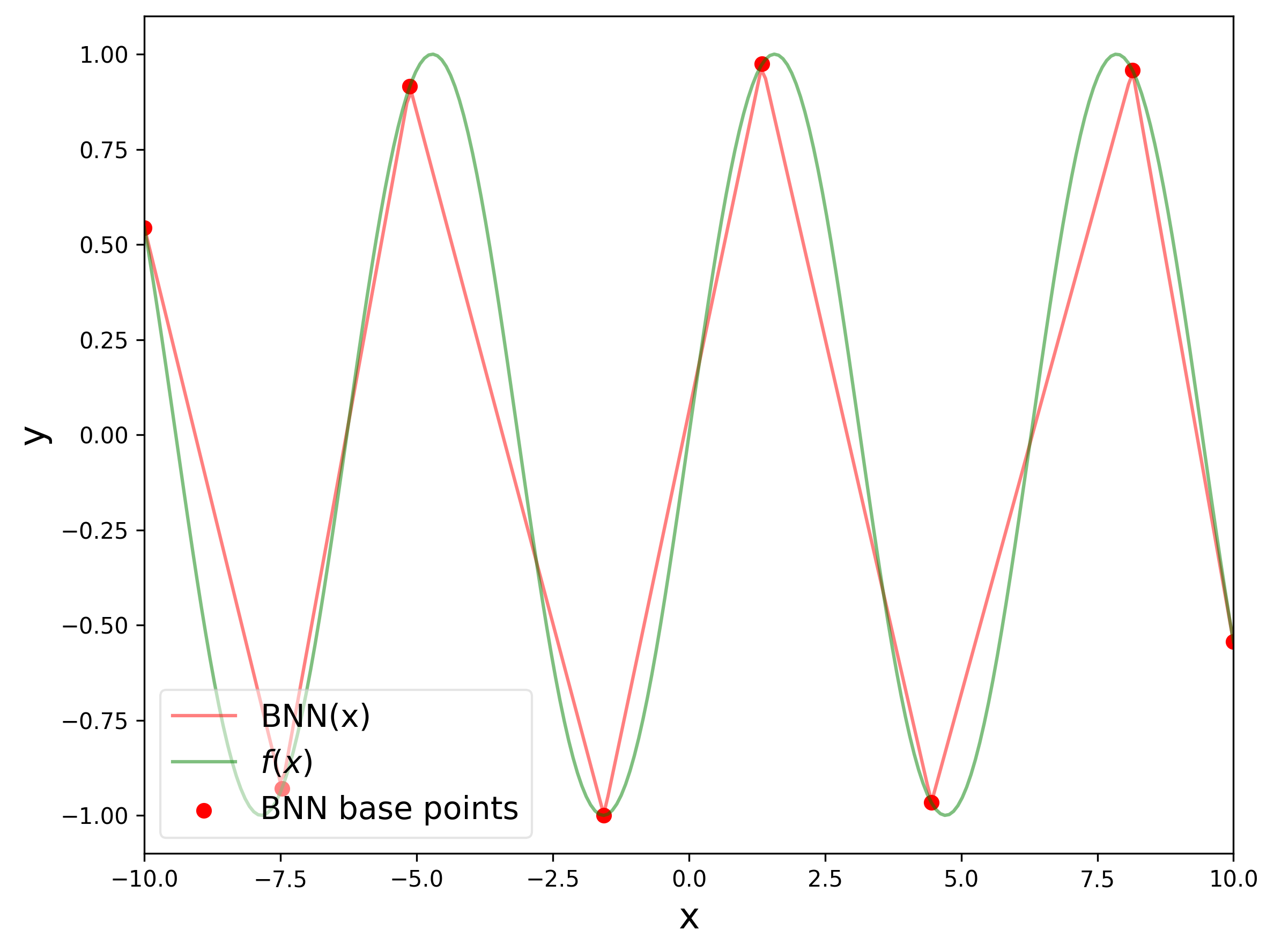}
    \caption{}
    \label{fig:comparingPEsAproxLWPE}
    \end{subfigure}
    \begin{subfigure}[b]{0.45\textwidth}
    \centering
    \includegraphics[width=\textwidth]{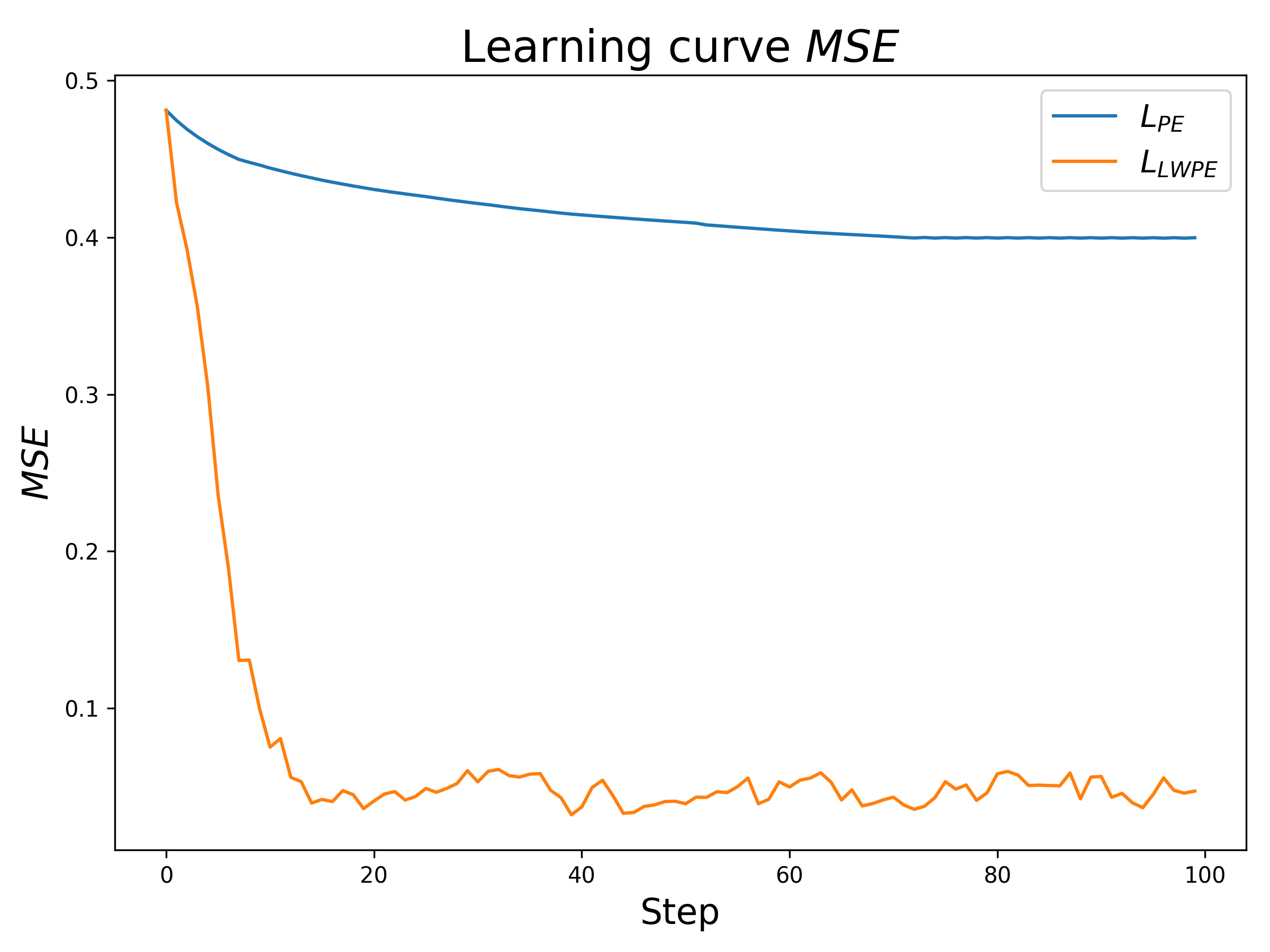}
    \caption{}
    \label{fig:comparingPEsLearningCurve}
    \end{subfigure}
    \caption{Comparison of function approximation using a $\BNN$ with 8 base points, optimized via two persistent entropy-based loss functions. (a) Initial state with randomly initialized base points. (b) 50th iteration, obtained using the standard persistent entropy-based loss $L_{\PE}$. (c) 50th iteration, obtained using the length-weighted persistent entropy-based loss $L_{\LWPE}$. (d) MSE learning curves showing convergence differences between $L_{\PE}$ and $L_{\LWPE}$.} \label{fig:comparingPEs}
\end{figure}

\subsection{$L_{\PE}$   vs $L_{\LWPE}$ topological loss}

Initially, we considered a topological loss based on the standard persistent entropy: \begin{equation*}
    L_{\PE} = |\PE_{\reff} - \PE_{\pred}|
\end{equation*}
where both $\PE_{\reff}$ and $\PE_{\pred}$ are the persistent entropy computed via the lower-star filtrations of the reference and predicted point clouds, respectively. The objective of this persistent entropy-based loss is to minimize the absolute difference between the persistent entropy derived from the reference and predicted point cloud, ensuring a similar topological structure. However, this standard persistent entropy-based approach presents a fundamental limitation, as discussed in Section \ref{sec:persistentEntropy}, since it does not necessarily guide the $\BNN$ and its base points towards an accurate pointwise approximation of the target function. This limitation is clearly shown in Fig \ref{fig:comparingPEs}, where, after optimizing the base points to approximate $f(x) = \sin(x)$ on the interval $[-10, 10]$ using $L_{\mathrm{PE}}$, the $\BNN$ converges to a representation that preserves the relative proportions of the peaks but fails to align with their true amplitudes and oscillatory structure. Consequently, even though the persistent entropy values of the reference and prediction remain close, the learned function does not capture the actual maxima and minima of the target.

To overcome this limitation, we employ our length-weighted persistent entropy ($\LWPE$), introduced in Section \ref{sec:persistentEntropy}, as the basis for the topological loss, replacing the standard persistent entropy. As shown in Fig.~\ref{fig:comparingPEs}, optimizing base points using $\LWPE$ loss instead of $\PE$ yields a significantly closer fit to the target function. This observation is further supported by the learning curves in Fig.\ref{fig:comparingPEsLearningCurve}: while the standard entropy variant results to slower and limited convergence, our length-weighted version achieves rapid and stable minimization of the MSE, indicating more effective learning dynamics. Both experiments used the stochastic gradient descent (SGD) optimization algorithm with a learning rate of 0.1, as well as limited resources: 50 epochs and 8 base points.

Consequently, we adopt $L_{\LWPE}$ as our default topological loss based on our proposed variant of persistent entropy throughout the remainder of the experiments. Formally, it is defined as:
\begin{equation*}
    L_{\LWPE} = |\LWPE_{\reff} - \LWPE_{\pred}|.
\end{equation*}

\begin{figure}[ht!]
\centering
\includegraphics[width=0.95\textwidth]{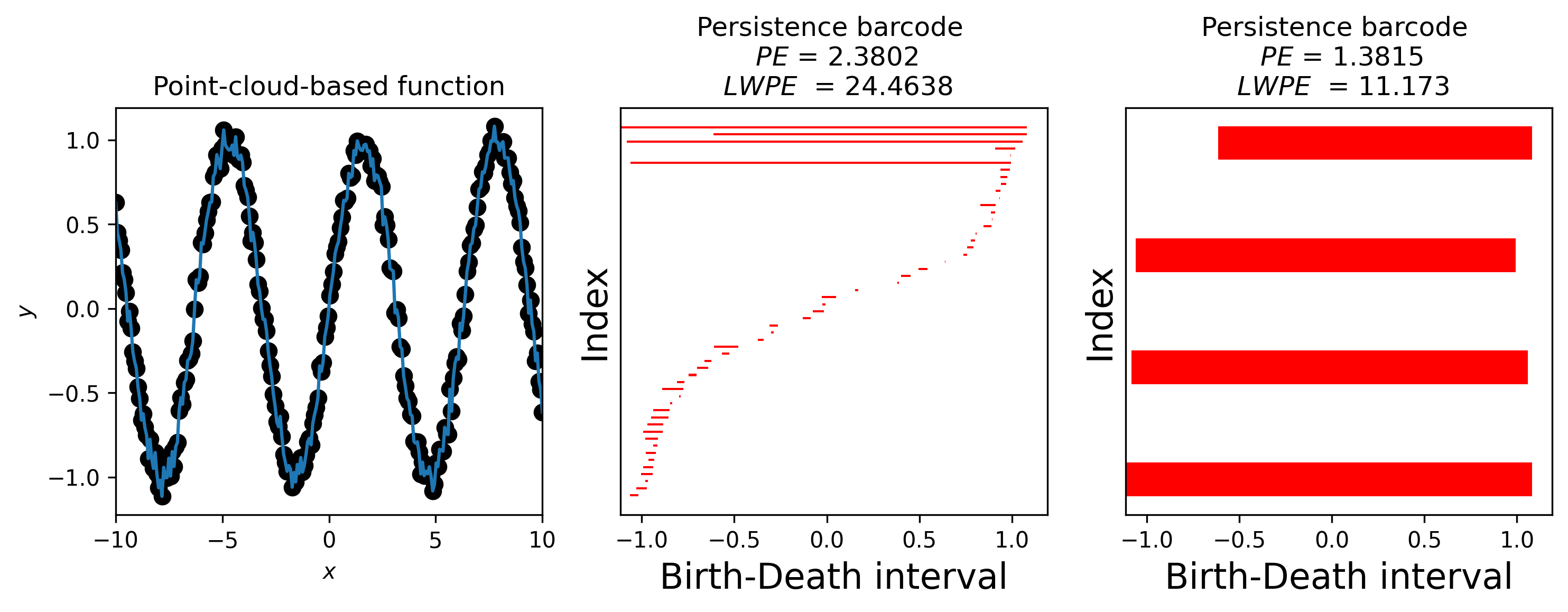}
\caption{Similar point-cloud-based function consisting of 250 points as in Fig. \ref{fig:entropyCalculation}, but with noise (left). The persistence barcode (middle) and its filtered version (right), showing only the 4 most significant bars, which illustrate the topological features that can be approximated using 8 base points to construct the $\BNN$.}
\label{fig:unknoiseandpb}
\end{figure}

\begin{figure}[ht!]
    \centering
    \begin{subfigure}[b]{0.3\textwidth}
    \centering
    \includegraphics[width=\textwidth]{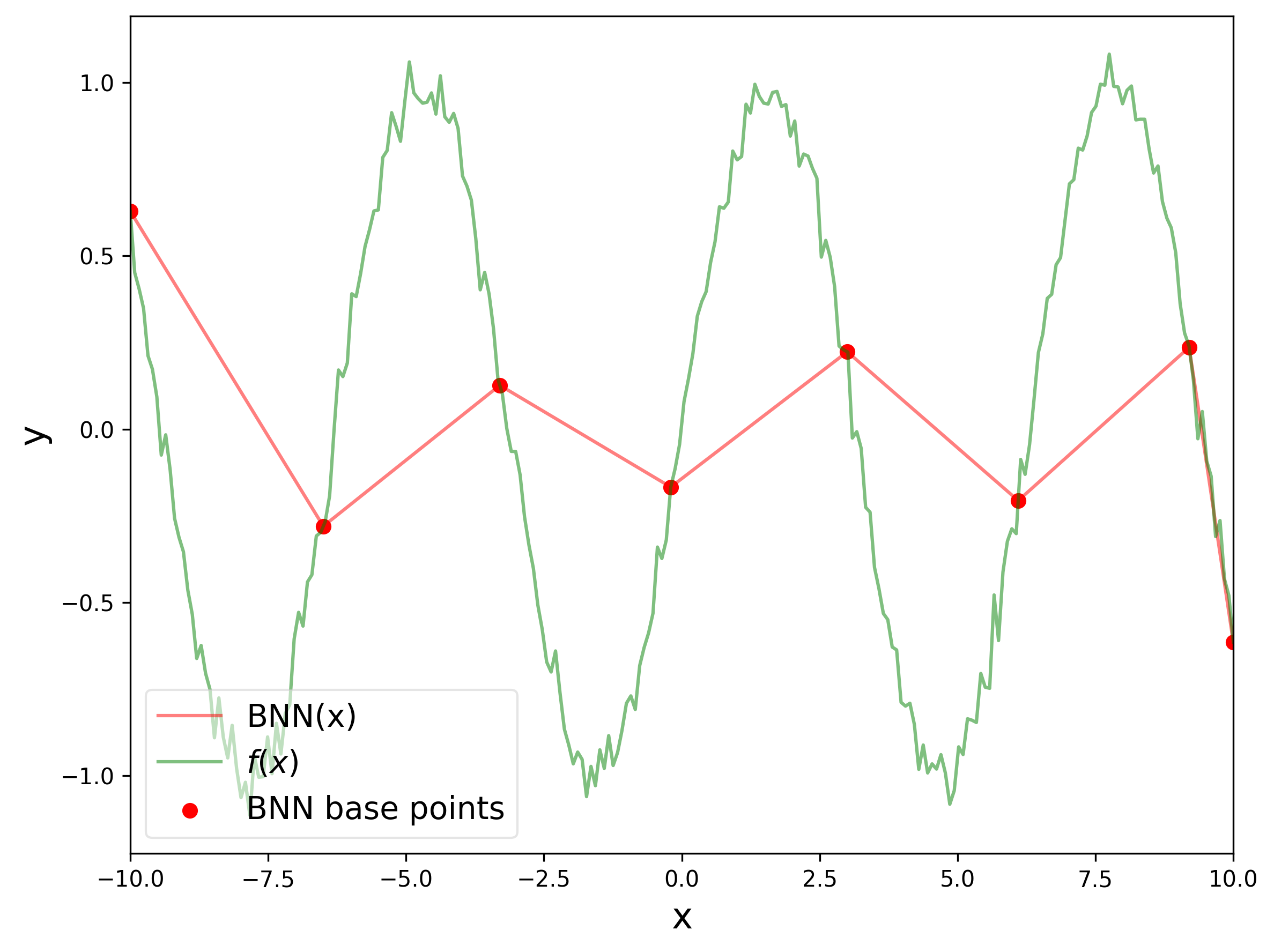}  
    \caption{}
    \end{subfigure}
    \begin{subfigure}[b]{0.3\textwidth}
    \centering
    \includegraphics[width=\textwidth]{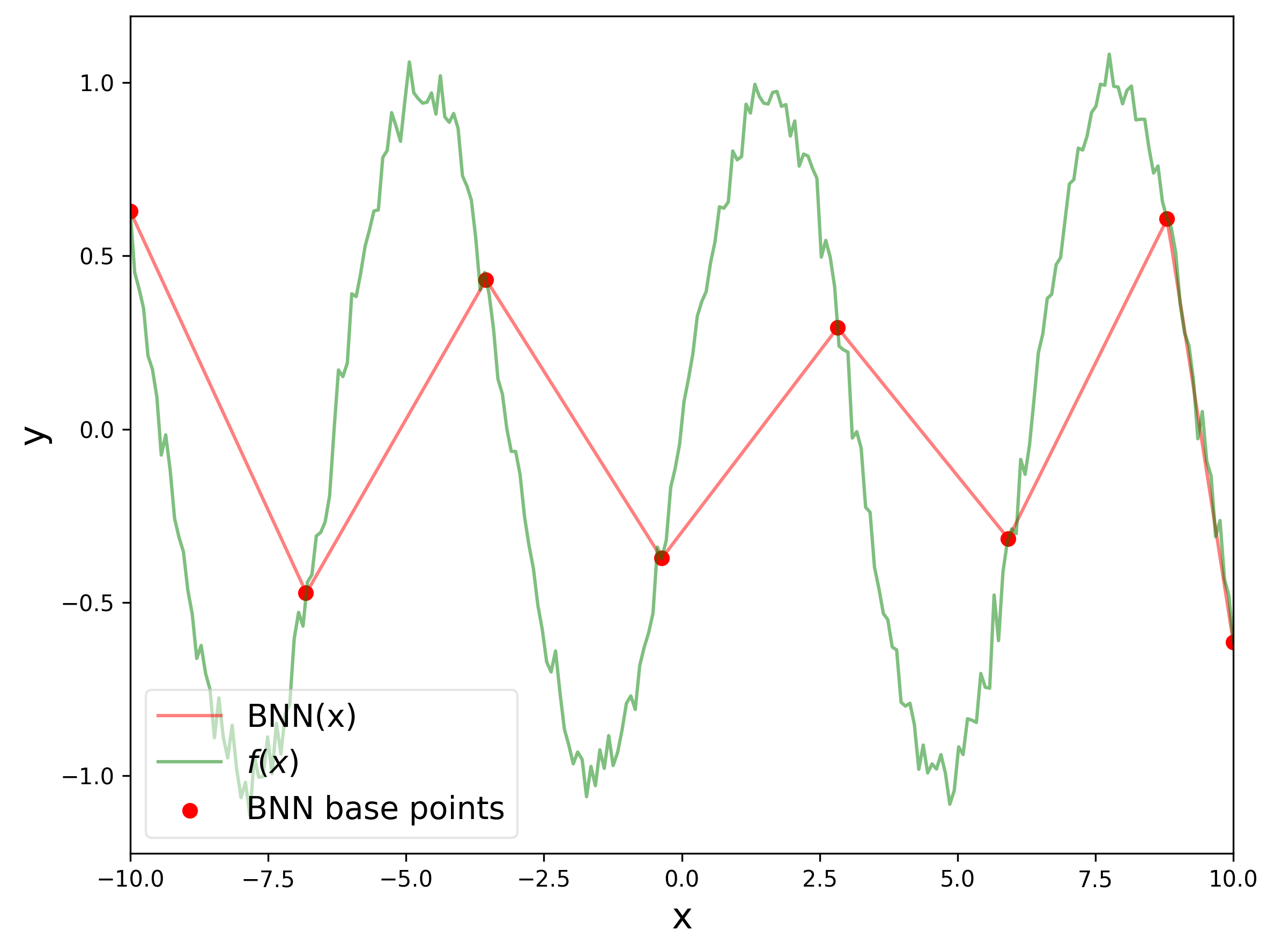}
    \caption{}
    \end{subfigure}
    \begin{subfigure}[b]{0.3\textwidth}
    \centering
    \includegraphics[width=\textwidth]{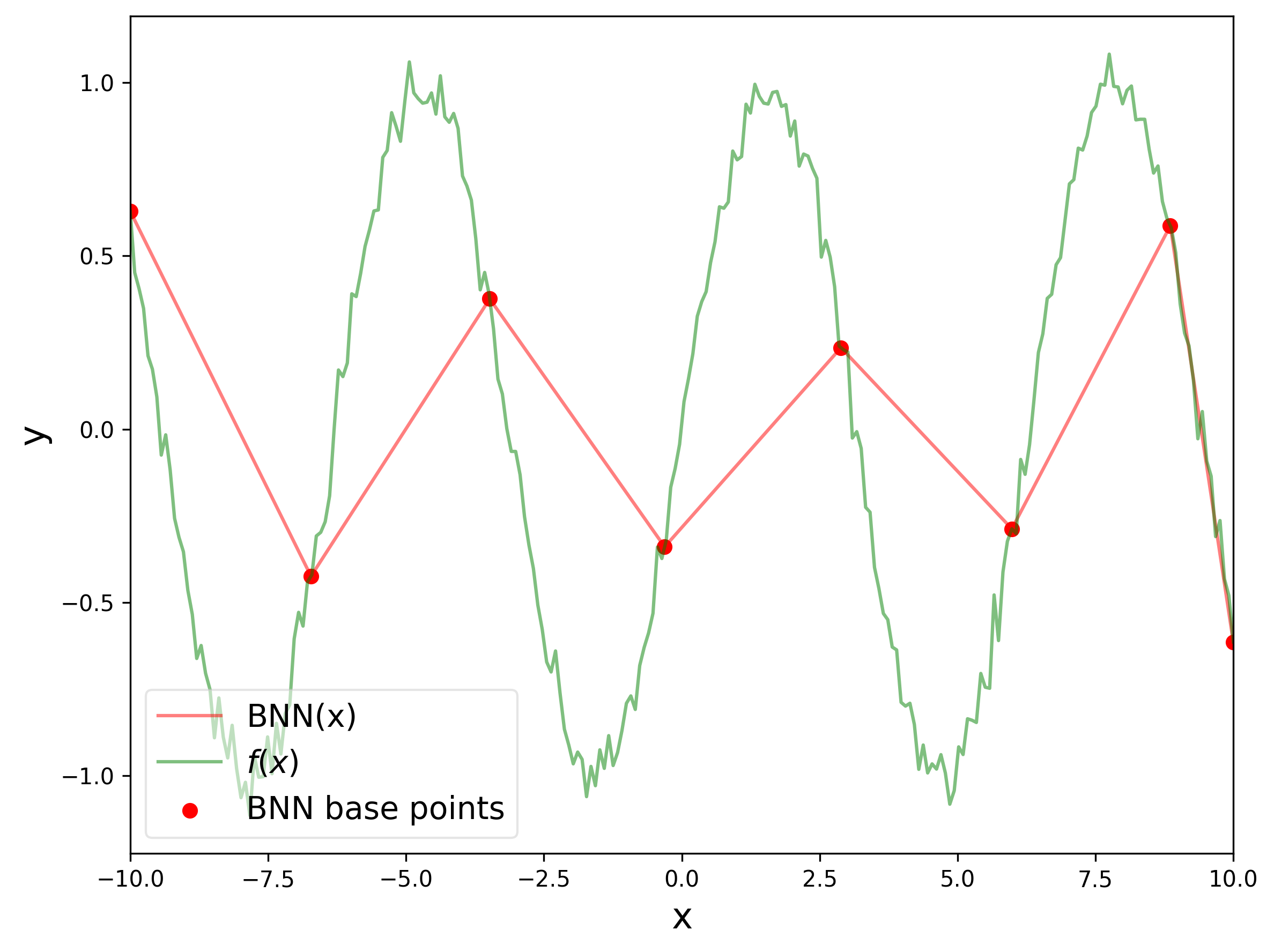}
    \caption{}
    \end{subfigure}
    \begin{subfigure}[b]{0.3\textwidth}
    \centering
    \includegraphics[width=\textwidth]{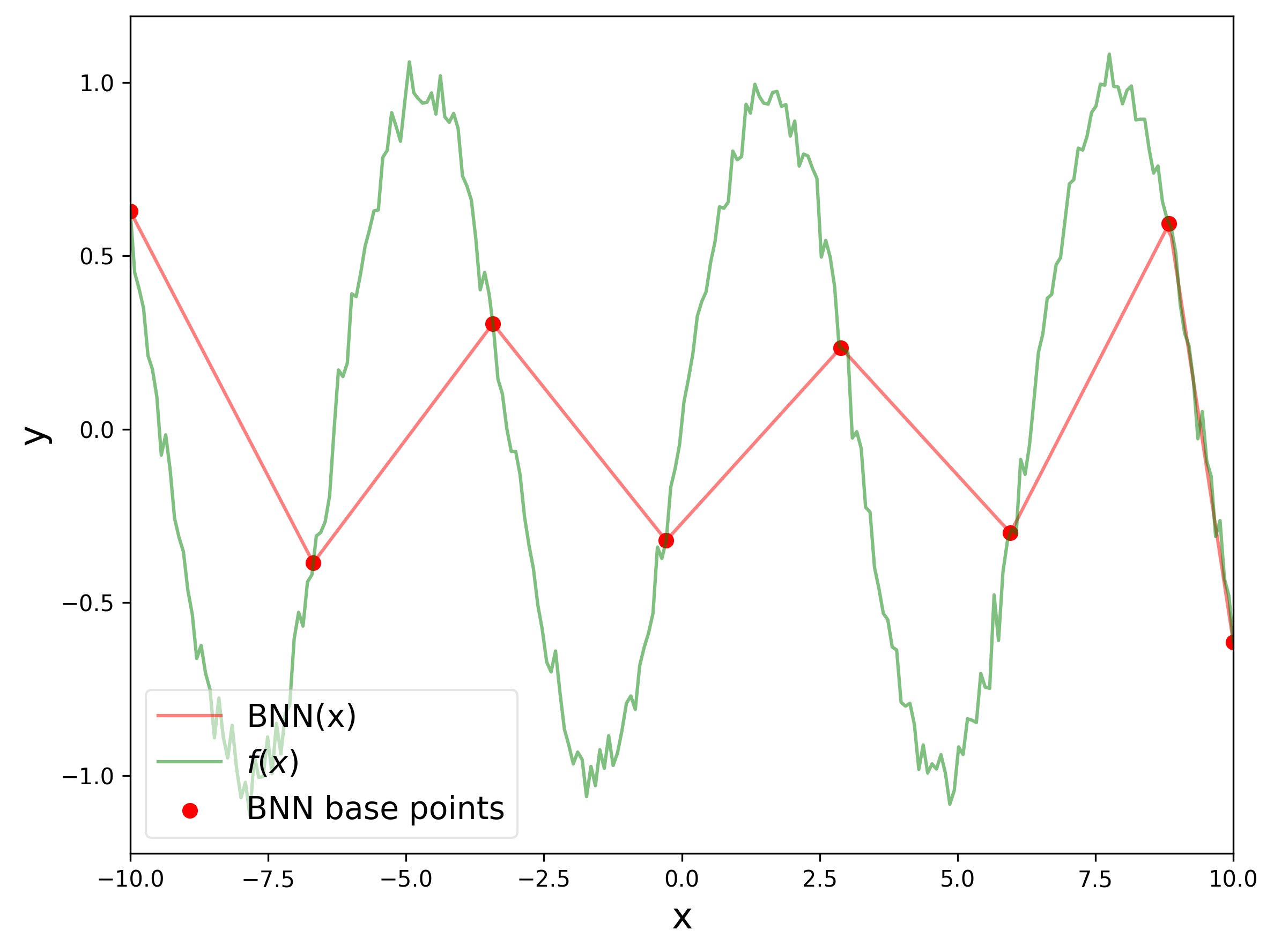}
    \caption{}
    \end{subfigure}
    \begin{subfigure}[b]{0.3\textwidth}
    \centering
    \includegraphics[width=\textwidth]{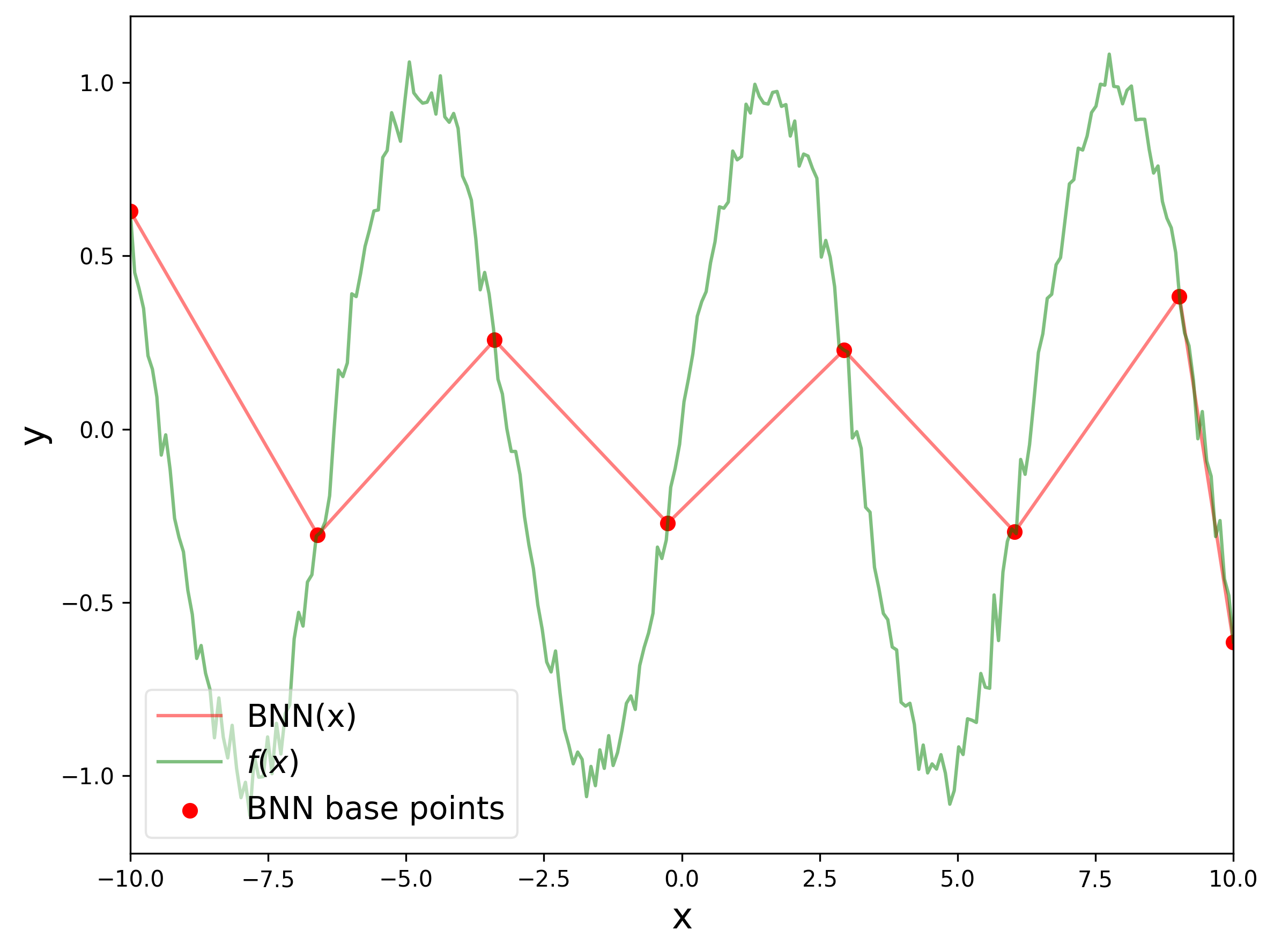}
    \caption{}
    \end{subfigure}
    \begin{subfigure}[b]{0.3\textwidth}
    \centering
    \includegraphics[width=\textwidth]{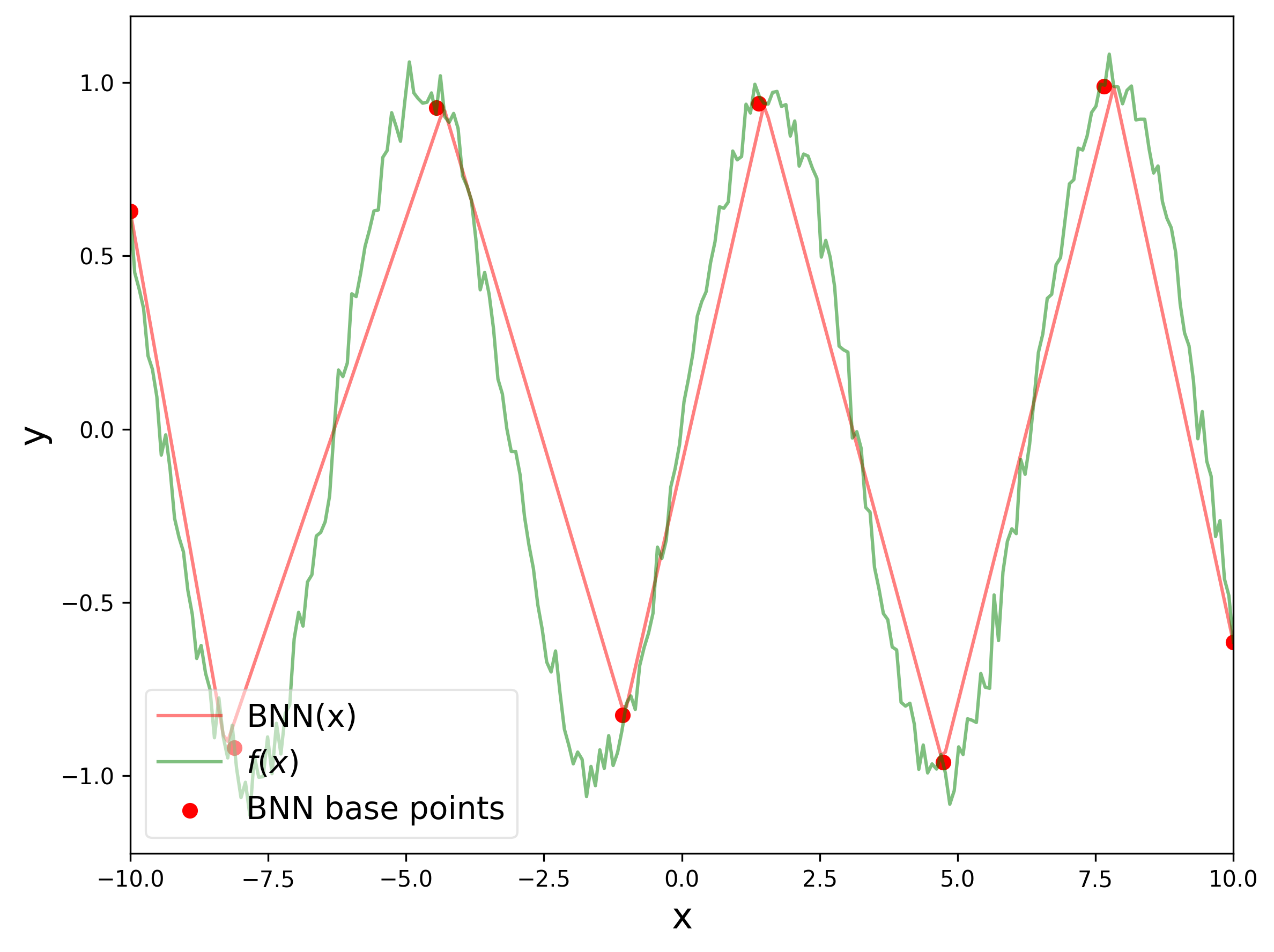}
    \caption{}
    \end{subfigure}
    \caption{Approximation of a noisy point-cloud-based function using a $\BNN$ with 8 base points, optimized via different loss functions. (a) Initial state with randomly placed base points. (b–f) Final epoch (50) BNN approximation using the following loss functions:(b) $\MSE$, (c) $\RMSE$, (d) $\MAE$, (e) LogCosh, (f) $L_{\LWPE}$.} \label{fig:aproxBNN_noise}
\end{figure}

\subsection{Robustness under noise and a limited number of base points}\label{subsec:noise}

To test the robustness of our framework, we evaluate its performance in approximating a noisy point cloud function using a limited number of base points. This scenario reflects more realistic applications, as in most real-world cases, point clouds will not be as smooth as the one presented previously.

First, as an illustrative example of Remark~\ref{re:constraint}, see Fig. \ref{fig:unknoiseandpb}, where we consider a point cloud $X$ with 250 points representing a noisy partial description of the sine function (left). As discussed above, and by analyzing the obtained persistence barcode computed from $X$, we can deduce that with $n = 8$ base points, we can 
approximate the sine function
with the $n/2 = 4$ longest bars, while treating the remaining bars of the barcode as noise (see the right plot of Fig. \ref{fig:unknoiseandpb}).

\begin{figure}[ht!]
    \centering
    \includegraphics[width=0.6\textwidth]{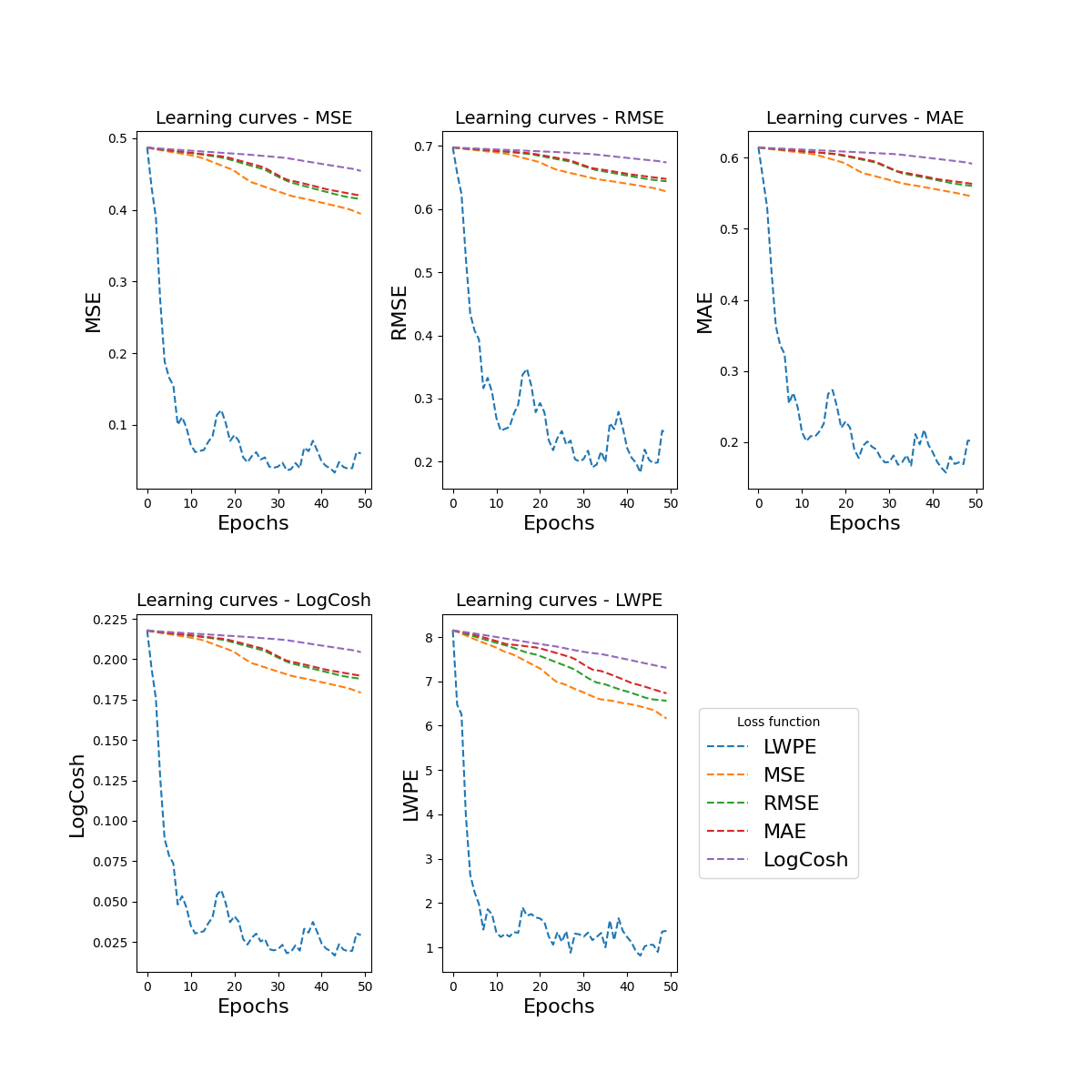}
    \caption{Learning curves for approximating a noisy point-cloud-based function using a $\BNN$ with 8 base points. The plots compare convergence across different loss functions used to optimize the base points: $\MSE$, $\RMSE$, $\MAE$, LogCosh, and our length-weighted persistent entropy loss ($L_{\LWPE}$). Curves are ordered row-wise: first row, first column is $\MSE$; first row, second column is RMSE; and so on.} \label{fig:learningcurves_aproxBNN_noise}
\end{figure}

Second, in Fig. \ref{fig:aproxBNN_noise}, we can see an example where we aim to approximate a noisy version of the sine function represented by a noisy point cloud, using limited resources (8 base points and 50 epochs), utilizing BNNs. We compared the results obtained with the $L_{\LWPE}$ topological loss against more commonly used loss functions, such as $\MSE$, $\RMSE$, $\MAE$, or the LogCosh. We have observed that with $L_{\LWPE}$, the $\BNN$ achieves a more effective distribution of base points and, consequently, a superior approximation under the same constraints. Furthermore, the learning curves in Fig. \ref{fig:learningcurves_aproxBNN_noise} show that optimizing the base points with the $L_{\LWPE}$ loss exhibits a superior and faster performance than using classical loss functions. This indicates that the $L_{\LWPE}$ loss provides more informative gradients during the early training stages, enabling a quicker and more accurate optimization of base point distributions, and consequently, a better approximation, reinforcing our framework's effectiveness in handling complex and noisy data, even under limited settings.

\subsection{Robustness under outliers}\label{subsec:outliers}

A further challenge in real-world data is the presence of outliers, i.e., extreme values that deviate significantly from the underlying structure of the data. To test the robustness of our framework under this setting, we consider again a noisy sine function (similar to Subsection~\ref{subsec:noise}), but we introduce some artificial outliers at some of its peaks. This creates a distorted point cloud where the global shape of the sine curve coexists with an anomalous observation that could potentially bias the learning process.

When trained exclusively with our topological loss based on $\LWPE$, the model tends to overfit to the outlier. This behavior is expected due to the sensitivity of $\LWPE$ to outliers: the outlier generates a highly persistent topological feature (e.g., a long bar in the persistence diagram), which dominates the $\LWPE$ computation and biases the optimization toward reproducing the anomalous value rather than the true underlying signal.

\begin{figure}[ht!]
    \centering
    \includegraphics[width=0.95\textwidth]{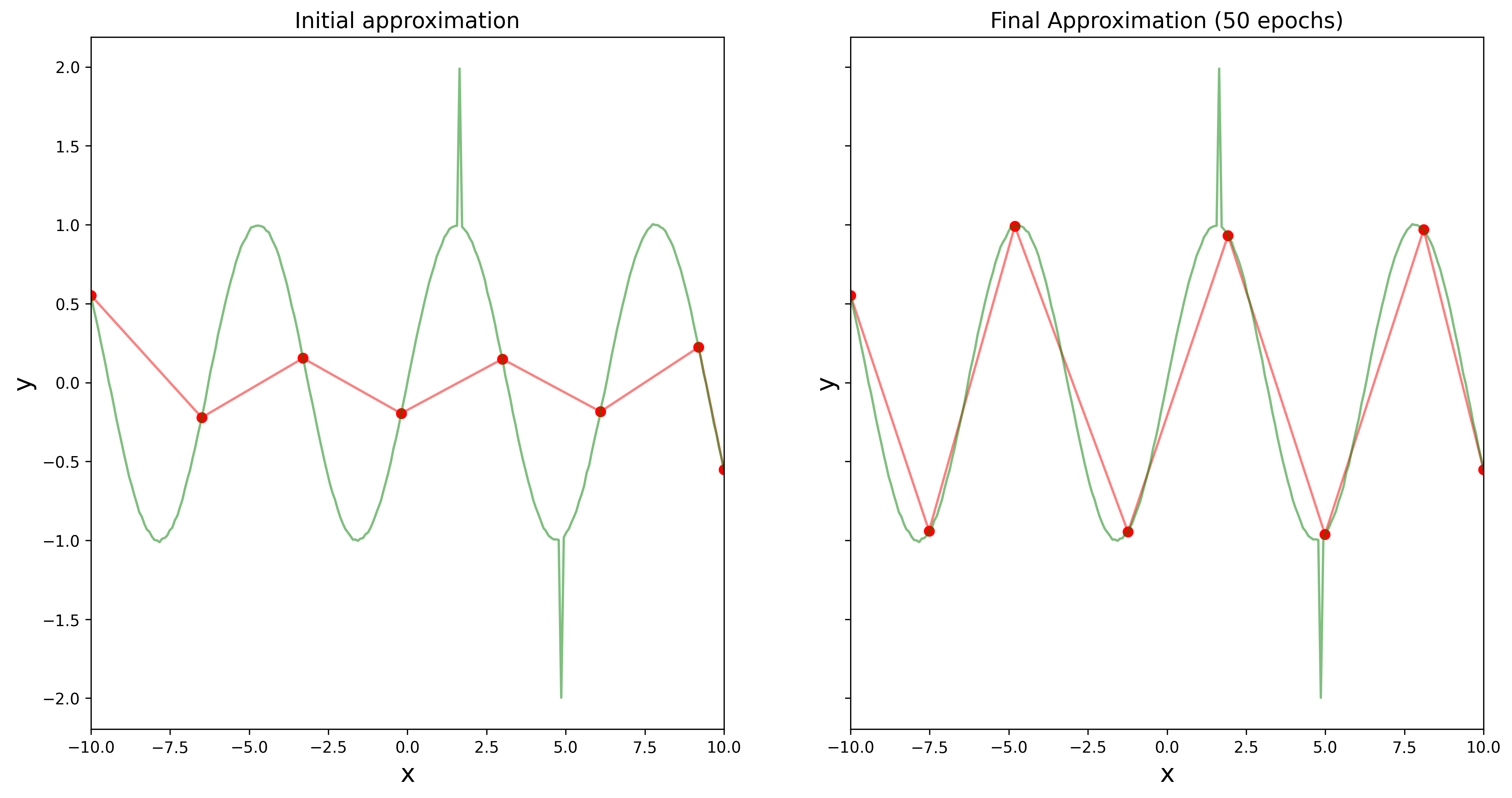}
    \caption{Approximation of a noisy point-cloud-based function containing outliers using a $\BNN$ with 8 base points, optimized using $L_{\LWPE}$ as a regularizer of MSE. On the left, the initial state with randomly placed base points, and on the right, the final BNN approximation after 50 epochs.} \label{fig:outlier_example}
\end{figure}

To address this limitation, we propose a hybrid training strategy that combines $\LWPE$ as a regularizer with a classical loss such as MSE. As shown in Fig.~\ref{fig:outlier_example}, this composite objective effectively balances local fidelity with global structural consistency. The result avoids two pitfalls: (1) the slow convergence and poor shape recovery observed when using MSE alone (as shown in Section~\ref{subsec:noise}), and (2) the outlier-driven distortion caused by relying solely on $\LWPE$. The hybrid approach thus achieves a high-quality approximation that respects both the fine-grained data and the overarching topology of the target function.

These findings underscore the robust performance even in the presence of data outliers, a common reality in practical applications. Moreover, the modular design of our framework allows the topological loss to be flexibly incorporated as a regularizer, enabling practitioners to tailor the trade-off between performance and structural integrity according to the demands of their specific domain.

\subsection{Testing our method with real-world data} 

\begin{figure}[ht!]
    \centering
    \includegraphics[width=0.6\textwidth]{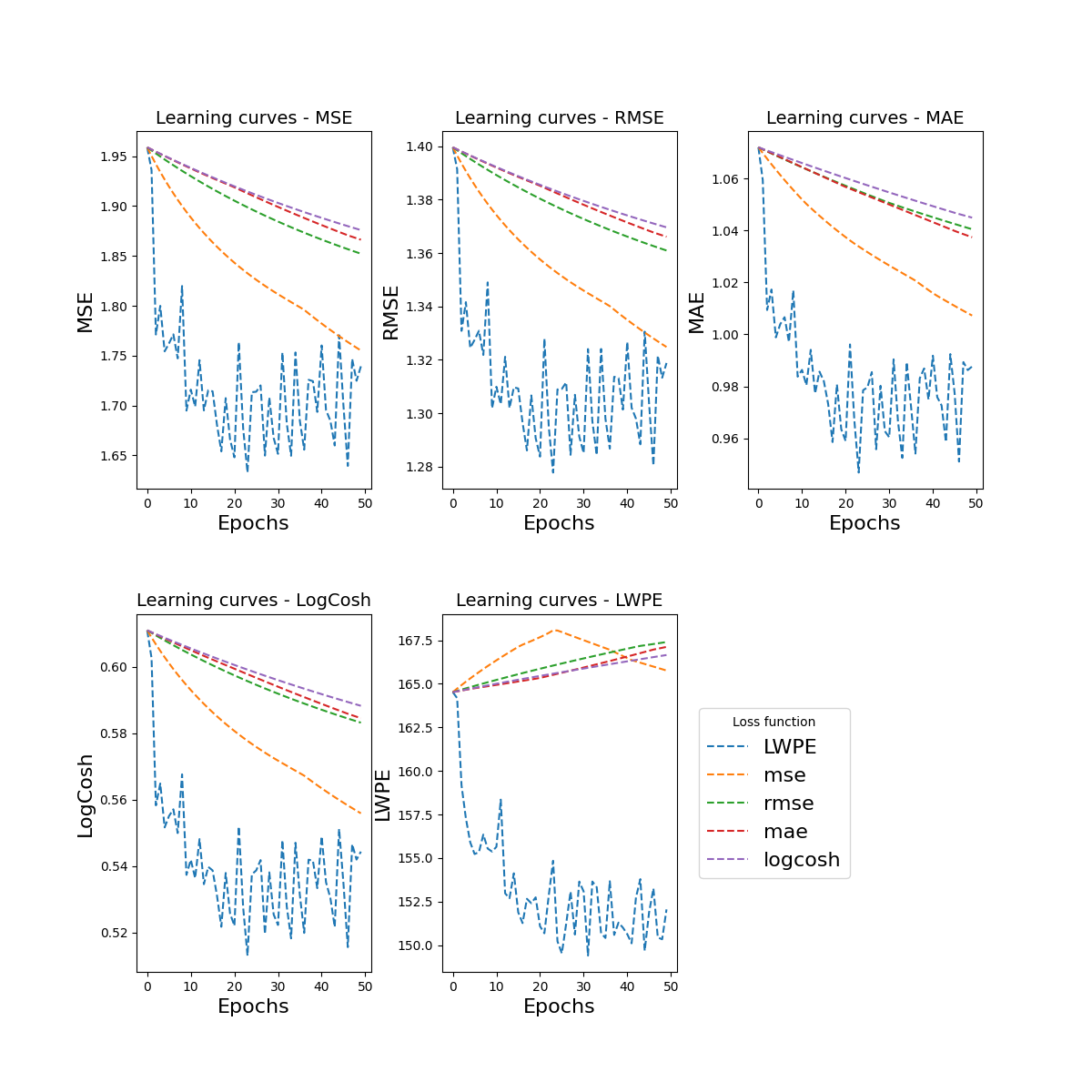}
    \caption{Learning curves for approximating gold ETF 2023 time series using a $\BNN$ with 30 base points. The plots compare convergence across different loss functions used to optimize the base points: $\MSE$, $\RMSE$, $\MAE$, LogCosh, and our length-weighted persistent entropy loss ($L_{\LWPE}$). Curves are ordered row-wise: first row, first column is $\MSE$; first row, second column is RMSE; and so on.} \label{fig:learningcurves_aproxBNN_Gold}
\end{figure}

To showcase the applicability of our method to real-world data, we approximate a financial time series: the daily price of a Gold ETF during the year 2023 (point-cloud-based function consisting of 365 points), obtained using the yfinance Python library\footnote{The yfinance documentation is available at \url{https://ranaroussi.github.io/yfinance/}}. Having a limited number of 30 base points, we aim to capture the signal’s essential structure, effectively treating it as a signal compression task that removes short-term noise. In Fig.~\ref{fig:learningcurves_aproxBNN_Gold}, we observe that $L_{\LWPE}$ again outperforms the classical loss functions, reducing approximation error across all metrics within the first epochs, highlighting its suitability for approximation tasks and time-series compression with a limited number of base points and epochs.

In summary, using the $L_{\LWPE}$ loss to optimize the base points that define $\BNN$s demonstrates clear benefits in approximating complex, noisy functions with very few base points and epochs.
This makes the approach particularly attractive for real-world applications with limited computational resources.
 
\section{Conclusions}\label{sec:conclusion}
In this work, we introduced a novel framework for function approximation that integrates geometric modeling with topological optimization. At the core of our approach lies the Barycentric Neural Network ($\BNN$), a shallow neural architecture whose structure and parameters are entirely determined by a set of trainable base points and their barycentric coordinates. This design enables exact representation of Continuous Piecewise Linear Functions ($\CPLF$s), and therefore, $\BNN$s serve as a flexible, interpretable, and structurally predefined architecture for approximating general continuous functions. Unlike traditional neural networks, which require optimization of internal weights, our $\BNN$ shifts the learning process to the optimization of the base points. To guide this optimization, we proposed a novel topological loss function based on the length-weighted persistent entropy. Our experimental results demonstrate that this $\BNN$-$\LWPE$ framework achieves superior performance for function approximation, including noisy functions, compared to using classical loss functions (MSE, RMSE, MAE, and LogCosh), especially under resource-constrained settings, such as a limited and fixed number of base points, and limited training epochs, making it a strong candidate for function approximation, as well as following with the principles of Green AI. In summary, our method provides a mathematically principled and computationally efficient alternative to standard deep learning approaches for function approximation, contributing to the development of more interpretable, sustainable, and theoretically grounded machine learning models.

\paragraph*{\textbf{Future work}}
Future research focuses on extending this approach to higher-dimensional function approximation and exploring its applications in more real-world scenarios. To obtain the simplicial complex $\K$ that covers a compact set $\C\subset\R^d$, we could, for example, compute Delaunay triangulations \cite{fortune2017voronoiDiagramsandDelaunayTriangulation} with a set of vertices $V$ being a set of points of $\R^d$. Although the time complexity of such construction is roughly $O(n^{\lfloor d/2\rfloor})$, it remains computationally feasible when $V$ is small enough (see \cite{Seidel1995}). In addition, we aim to investigate adaptive initialization schemes for the base points, possibly incorporating prior information or combining topological and statistical heuristics. Such approaches could enhance robustness and reduce dependence on careful manual initialization. Notably, our experiments already demonstrate that, whether base points are initialized uniformly or randomly, our topological loss consistently yields a better optimization process compared to classical losses. Furthermore, efforts will be directed toward formally proving the advantages of using the topological loss $\LWPE$. Additionally, the development of hybrid loss functions, which combine topological descriptors with classical metrics, may further enhance optimization in settings with noisy or weakly structured data, or containing outliers, as we checked in Subsection~\ref{subsec:outliers} where we show how our method can be adapted to be robust to outliers. Finally, while the current BNN architecture is designed to exactly represent CPLFs, future work will explore extending this framework to approximate smoother functions by allowing nonlinear interpolations between base points. This generalization would enable the model to capture higher-order smoothness while maintaining its geometric interpretability and parameter efficiency. Moreover, such an approach could naturally generalize to multivariate settings, broadening the applicability of the proposed method.

\paragraph*{\textbf{Code Availability}}
All code for the proposed methodology, as well as for generate the results presented in this manuscript, are publicly available in a Github repository \footnote{\url{https://github.com/victosdur/Function_Approximation_Using_BNN_and_LWPE.git}}.

%\paragraph*{\textbf{Acknowledgements}} 
%We want to thank the reviewers, who gave us useful comments to improve the content of this paper, as well as ideas for future work. 
%This work was partially supported by REXASI-PRO H-EU project, call HORIZON-CL4-2021-HUMAN-01-01, Grant agreement ID: 101070028.

\paragraph*{\textbf{Disclosure of Interest}}
The authors have no competing interests to declare that are relevant to the content of this article.

% It is now necessary to declare any competing interests or to specifically
% state that the authors have no competing interests. Please place the
% statement with a bold run-in heading in small font size beneath the
% (optional) acknowledgments\footnote{If EquinOCS, our proceedings submission
% system, is used, then the disclaimer can be provided directly in the system.},
% for example: The authors have no competing interests to declare that are
% relevant to the content of this article. Or: Author A has received research
% grants from Company W. Author B has received a speaker honorarium from
% Company X and owns stock in Company Y. Author C is a member of committee Z.

\bibliographystyle{plain}
\bibliography{biblio}
% \nocite{*}

\end{document}